\documentclass[10pt,twocolumn,letterpaper]{article}

\usepackage{cvpr}
\usepackage{times}
\usepackage{epsfig}
\usepackage{graphicx}
\usepackage{amsmath}
\usepackage{amssymb}
\usepackage{multirow}
\usepackage{enumitem}% http://ctan.org/pkg/enumitem

\usepackage[pagebackref=true,breaklinks=true,letterpaper=true,colorlinks,bookmarks=false]{hyperref}

\cvprfinalcopy % *** Uncomment this line for the final submission

\def\cvprPaperID{784} % *** Enter the CVPR Paper ID here

\ifcvprfinal\fi
\begin{document}

%%%%%%%%% TITLE
\title{Deep View Morphing}

\author{Dinghuang Ji\thanks{The majoriy of the work has been done during this author's 2016 summer internship at Ricoh Innovations.}\\
UNC at Chapel Hill\\
{\tt\small jdh@cs.unc.edu}
% For a paper whose authors are all at the same institution,
% omit the following lines up until the closing ``}''.
% Additional authors and addresses can be added with ``\and'',
% just like the second author.
% To save space, use either the email address or home page, not both
\and
Junghyun Kwon\thanks{This author is currently at SAIC Innovation Center.}\\
Ricoh Innovations\\
{\tt\small junghyunkwon@gmail.com}
\and
Max McFarland\\
Ricoh Innovations\\
{\tt\small max@ric.ricoh.com}
\and 
Silvio Savarese\\
Stanford University\\
{\tt\small ssilvio@stanford.edu}
}
 
\maketitle
%\thispagestyle{empty}

%%%%%%%%% ABSTRACT
\begin{abstract}
Recently, convolutional neural networks (CNN) have been successfully applied to view synthesis problems. However, such CNN-based methods can suffer from lack of texture details, shape distortions, or high computational complexity. In this paper, we propose a novel CNN architecture for view synthesis called ``Deep View Morphing" that does not suffer from these issues. To synthesize a middle view of two input images, a rectification network first rectifies the two input images. An encoder-decoder network then generates dense correspondences between the rectified images and blending masks to predict the visibility of pixels of the rectified images in the middle view. A view morphing network finally synthesizes the middle view using the dense correspondences and blending masks. We experimentally show the proposed method significantly outperforms the state-of-the-art CNN-based view synthesis method.  
\end{abstract}

%%%%%%%%% INTRODUCTION
\section{Introduction}
View synthesis is to create unseen novel views based on a set of available existing views. It has many appealing applications in computer vision and graphics such as virtual 3D tour from 2D images and photo editing with 3D object manipulation capabilities. Traditionally, view synthesis has been solved by image-based rendering  \cite{Beier_92_pami,Jones_95_pami,Katayama_95_spie,Shashua_95_pami,Levoy_96_siggraph,Gortler_96_siggraph,Seitz_1996_siggraph,Vetter_97_pami} 
 and 3D model-based rendering \cite{Horry,Oh,Zhang,Hoiem,Furukawa_2010_cvpr,Wu_3dv_2013,Jared_2015_cvpr} . 
 
Recently, convolutional neural networks (CNN) have been successfully applied to various view synthesis problems, {\em e.g.}, multi-view synthesis from a single view \cite{Yang,tatarchenko2016multi}, view interpolation \cite{Flynn_15_iccv}, or both \cite{Zhou_eccv}. While their results are impressive and promising, they still have limitations. Direct pixel generation methods such as \cite{Yang} and \cite{tatarchenko2016multi} have a main advantage that the overall geometric shapes are well predicted but their synthesis results usually lack detailed textures. On the other hand, the pixel sampling methods such as \cite{Flynn_15_iccv} and \cite{Zhou_eccv} can synthesize novel views with detailed textures but they suffer from high computational complexity \cite{Flynn_15_iccv} or geometric shape distortions \cite{Zhou_eccv}. 

\begin{figure}
\centering
\includegraphics[width=3.3in]{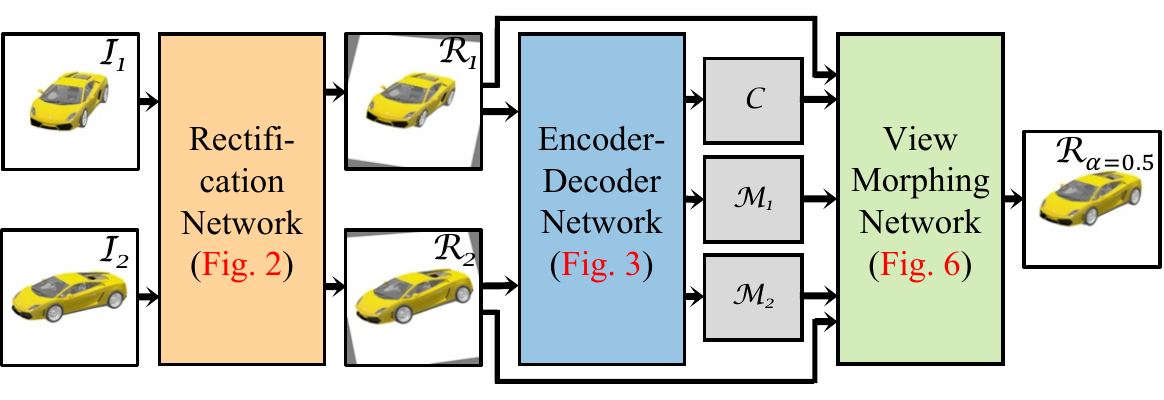}
\caption{Overall pipeline of Deep View Morphing. A rectification network (orange, Section~\ref{sec:rectification_network}) takes in $\mathcal{I}_1$ and $\mathcal{I}_2$ and outputs a rectified pair $\mathcal{R}_1$ and $\mathcal{R}_2$. Then an encoder-decoder network (blue, Section~\ref{sec:encoder-decoder}) takes in $\mathcal{R}_1$ and $\mathcal{R}_2$ and outputs the dense correspondences $\mathcal{C}$ and blending masks $\mathcal{M}_1$ and $\mathcal{M}_2$. Finally, a view morphing network(green, Section~ \ref{sec:view_morphing_network}) synthesizes a middle view $\mathcal{R}_{\alpha=0.5}$ from $\mathcal{R}_1$, $\mathcal{R}_2$, $\mathcal{M}_1$, $\mathcal{M}_2$, and $\mathcal{C}$.}
\label{fig:whole_pipeline}
\end{figure} 

In this paper, we propose a novel CNN architecture that can efficiently synthesize novel views with detailed textures as well as well-preserved geometric shapes under the view interpolation setting . We are mainly inspired by View Morphing, the classic work by Seitz and Dyer \cite{Seitz_1996_siggraph}, which showed it is possible to synthesize shape-preserving novel views by simple linear interpolation of the corresponding pixels of a rectified image pair. Following the spirit of View Morphing, our approach introduces a novel deep CNN architecture to generalize the procedure in \cite{Seitz_1996_siggraph}---for that reason, we named it Deep View Morphing (DVM). 

Figure~\ref{fig:whole_pipeline} shows the overall pipeline of DVM. A rectification network (orange in Fig.~\ref{fig:whole_pipeline}) takes in a pair of input images and outputs a rectified pair. Then an encoder-decoder network (blue in Fig.~\ref{fig:whole_pipeline}) takes in the rectified pair and outputs dense correspondences between them and blending masks. A view morphing network (green in Fig.~\ref{fig:whole_pipeline}) finally synthesizes a middle view using the dense correspondences and blending masks. The novel aspects of DVM are: 
\begin{itemize}[noitemsep,topsep=0pt,leftmargin=15pt]
\item The idea of adding a rectification network before the view synthesis phase---this is critical in that rectification guarantees the correspondences should be 1D, which makes the correspondence search by the encoder-decoder network significantly easier. As a result, we can obtain highly accurate correspondences and consequently high quality view synthesis results. The rectification network  is inspired by \cite{STN}, which learns how to transform input images to maximize the classification accuracy. In DVM, the rectification network learns how to transform an input image pair for rectification.
\item DVM does not require additional information other than the input image pair compared to \cite{Zhou_eccv} that needs viewpoint transformation information and \cite{Flynn_15_iccv} that needs camera parameters and higher-dimensional intermediate representation of input images.
\item As all layers of DVM are differentiable, it can be efficiently trained end-to-end with a single loss at the end.
\end{itemize}

In Section~\ref{sec:exp}, we experimentally show that: (i) DVM can produce high quality view synthesis results not only for synthesized images rendered with ShapeNet 3D models \cite{shapenet} but also for real images of Multi-PIE data \cite{MultiPIE}; (ii) DVM substantially outperforms \cite{Zhou_eccv}, the state-of-the-art CNN-based view synthesis method under the view interpolation setting, via extensive qualitative and quantitative comparisons; (iii) DVM generalizes well to categories not used in training; and (iv) all intermediate views beyond the middle view can be synthesized utilizing the predicted correspondences.

%%%%%%%%% RELATED WORK
\subsection{Related works}
{\setlength{\parindent}{0cm}{\bf View synthesis by traditional methods.}} Earlier view synthesis works based on image-based rendering include the well-known Beier and Neely's feature-based morphing \cite{Beier_92_pami} and learning-based methods to produce novel views of human faces \cite{Vetter_97_pami} and human stick figures \cite{Jones_95_pami}. For shape-preserving view synthesis, geometric constraints have been added such as known depth values at each pixel \cite{Chen_93_siggraph}, epipolar constraints between a pair of images \cite{Seitz_1996_siggraph}, and trilinear tensors that link correspondences between triplets of images \cite{Shashua_95_pami}. In this paper, DVM generalizes the procedure in \cite{Seitz_1996_siggraph} using a single CNN architecture.

Structure-from-motion can be used for view synthesis by rendering reconstructed 3D models onto virtual views. This typically involves the steps of camera pose estimation \cite{Jared_2015_cvpr,Wu_3dv_2013,zheng_iccv2015} and image based 3D reconstruction \cite{Furukawa_2010_cvpr,zheng_cvpr2014}. However, as these methods reply on pixel correspondences across views, their results can be problematic for textureless regions. The intervention of users is often required to obtain accurate 3D geometries of objects or scenes \cite{Horry,Oh,Zhang,Hoiem}. Compared to these 3D model-based methods, DVM can predict highly accurate correspondences even for textureless regions and does not need the intervention of users or domain experts.\\

{\setlength{\parindent}{0cm}{\bf View synthesis by CNN.}} Hinton et al. \cite{Hinton} proposed auto-encoder architectures to learn a group of auto-encoders that learn how to geometrically transform input images. Dosovitiskiy et al. \cite{Dosovitiskiy} proposed a generative CNN architecture to synthesize images given the object identity and pose. Yang et al. \cite{Yang} proposed recurrent convolutional encoder-decoder networks to learn how to synthesize images of rotated objects from a single input image by decoupling pose and identity latent factors while Tatarchenko et al. \cite{tatarchenko2016multi} proposed a similar CNN architecture without explicit decoupling of such factors. A key limitation of \cite{Dosovitiskiy,Yang,tatarchenko2016multi} is output images are often blurry and lack detailed textures as they generate pixel values from scratch. In order to solve this issue, Zhou et al. \cite{Zhou_eccv} proposed to sample from input images by predicting the appearance flow between the input and output for both multi-view synthesis from a single view and view interpolation. To resolve disocclusion and geometric distortion, Park et al. \cite{tvsn_cvpr2017} further proposed disocclusion aware flow prediction followed by image completion and refinement stage. Flynn et al. \cite{Flynn_15_iccv} also proposed to optimally sample and blend from plane sweep volumes created from input images for view interpolation. 

Among these CNN-based view synthesis methods, \cite{Flynn_15_iccv} and \cite{Zhou_eccv} are closely related to DVM as they can solve the view interpolation problem. Both demonstrated impressive view interpolation results, but they still have limitations. Those related to \cite{Flynn_15_iccv} include: (i) the need of creating plane sweep volumes, (ii) higher computational complexity, and (iii) assumption that camera parameters are known in testing. Although \cite{Zhou_eccv} is computationally more efficient than \cite{Flynn_15_iccv} and does not require known camera parameters in testing, it still has some limitations. For instance, \cite{Zhou_eccv}  assumes that viewpoint transformation is given in testing. Moreover, lack of geometric constraints on the appearance flow can lead to shape or texture distortions. Contrarily, DVM can synthesize novel views efficiently without the need of any additional information other than two input images. Moreover, the rectification of two input images in DVM plays a key role in that it imposes geometric constraints that lead to shape-preserving view synthesis results. 

%================================================%
%                          Classical View Morphing                                    %
%================================================%
%%%%%%%%% CLASSICAL VIEW MORHPING
\section{View Morphing}
We start with briefly summarizing View Morphing \cite{Seitz_1996_siggraph} for the case of unknown camera parameters.

\subsection{Rectification}
Given two input images $\mathcal{I}_1$ and $\mathcal{I}_2$, the first step of View Morphing is to rectify them by applying homographies to each of them to make the corresponding points appear on the same rows. Such homographies can be computed from the fundamental matrix \cite{Hartley}. The rectified image pair can be considered as captured from two parallel view cameras. In \cite{Seitz_1996_siggraph}, it is shown that the linear interpolation of parallel views yields shape-preserving view synthesis results.

\subsection{View synthesis by interpolation}
Let $\mathcal{R}_1$ and $\mathcal{R}_2$ denote the rectified versions of $\mathcal{I}_1$ and $\mathcal{I}_2$. Novel view images can be synthesized by linearly interpolating positions and colors of corresponding pixels of $\mathcal{R}_1$ and $\mathcal{R}_2$. As the image pair is already rectified, such synthesis can be done on a row by row basis . 

Let $P_1 = \{p_1^{1}, \ldots, p_1^{N}\}$ and  $P_2 = \{p_2^{1}, \ldots, p_2^{N}\}$ denote the point correspondence sets between $\mathcal{R}_1$ and $\mathcal{R}_2$ where $p_1^i, p_2^j \in \Re^2$ are corresponding points when $i = j$. With $\alpha$ between 0 and 1, a novel view  $\mathcal{R}_\alpha$ can be synthesized as
\begin{equation}
\mathcal{R}_\alpha \left((1 - \alpha) p_1^i + \alpha p_2^i \right) = (1-\alpha)\mathcal{R}_1( p_1^i ) + \alpha\mathcal{R}_2 ( p_2^i ),
\label{eqn:interpolation}
\end{equation}
where $i=1,\ldots,N$.  As point correspondences found by feature matching are usually sparse, more correspondences need to be determined by interpolating the existing ones. Extra steps are usually further applied to deal with folds or holes caused by the visibility changes between $\mathcal{R}_1$ and $\mathcal{R}_2$.

\subsection{Post-warping}
\label{sec:post-warping}
As $\mathcal{R}_\alpha$ is synthesized on the image plane determined by the image planes of the rectified pair $\mathcal{R}_1$ and $\mathcal{R}_2$, it might not represent desired views. Therefore, post-warping with homographies can be optionally applied to $\mathcal{R}_\alpha$ to obtain desired views. Such homographies can be determined by user-specified control points.

%================================================%
%                             Deep View Morphing                                      %
%================================================%
\section{Deep View Morphing}
\label{sec:deep_view_morphing}

DVM is an end-to-end generalization of View Morphing by a single CNN architecture shown in Fig.~\ref{fig:whole_pipeline}. The rectification network (orange in Fig.~\ref{fig:whole_pipeline}) first rectifies two input images $\mathcal{I}_1$ and $\mathcal{I}_2$ without the need of having point correspondences across views. The encoder-decoder network (blue in Fig.~\ref{fig:whole_pipeline}) then outputs the dense correspondences $\mathcal{C}$ between the rectified pair $\mathcal{R}_1$ and $\mathcal{R}_2$ and blending masks $\mathcal{M}_1$ and $\mathcal{M}_2$. Finally, the view morphing network (green in Fig.~\ref{fig:whole_pipeline}) synthesizes a novel view $\mathcal{R}_{\alpha=0.5}$ from $\mathcal{R}_1$, $\mathcal{R}_2$, $\mathcal{M}_1$, $\mathcal{M}_2$, and $\mathcal{C}$. All layers of DVM are differentiable and it allows efficient end-to-end training. Although DVM is specifically configured to synthesize the middle view of $\mathcal{R}_1$ and $\mathcal{R}_2$, we can still synthesize all intermediate views utilizing the predicted dense correspondences as shown in Fig.~\ref{fig:interpolation_results}.

What is common between the rectification network and encoder-decoder network is they require a mechanism to encode correlations between two images as a form of CNN features. Similarly to \cite{dosovitskiy2015flownet}, we can consider two possible ways of such mechanisms: (i) early fusion by channel-wise concatenation of raw input images and (ii) late fusion by channel-wise concatenation of CNN features of input images. We chose to use the early fusion for the rectification network and late fusion for the encoder-decoder network (see Appendix A for in-depth analysis). We now present the details of each sub-network. 

%%%%%%%%% Rectification Network
\subsection{Rectification network}
\label{sec:rectification_network}

\begin{figure}
\centering
\includegraphics[width=3in]{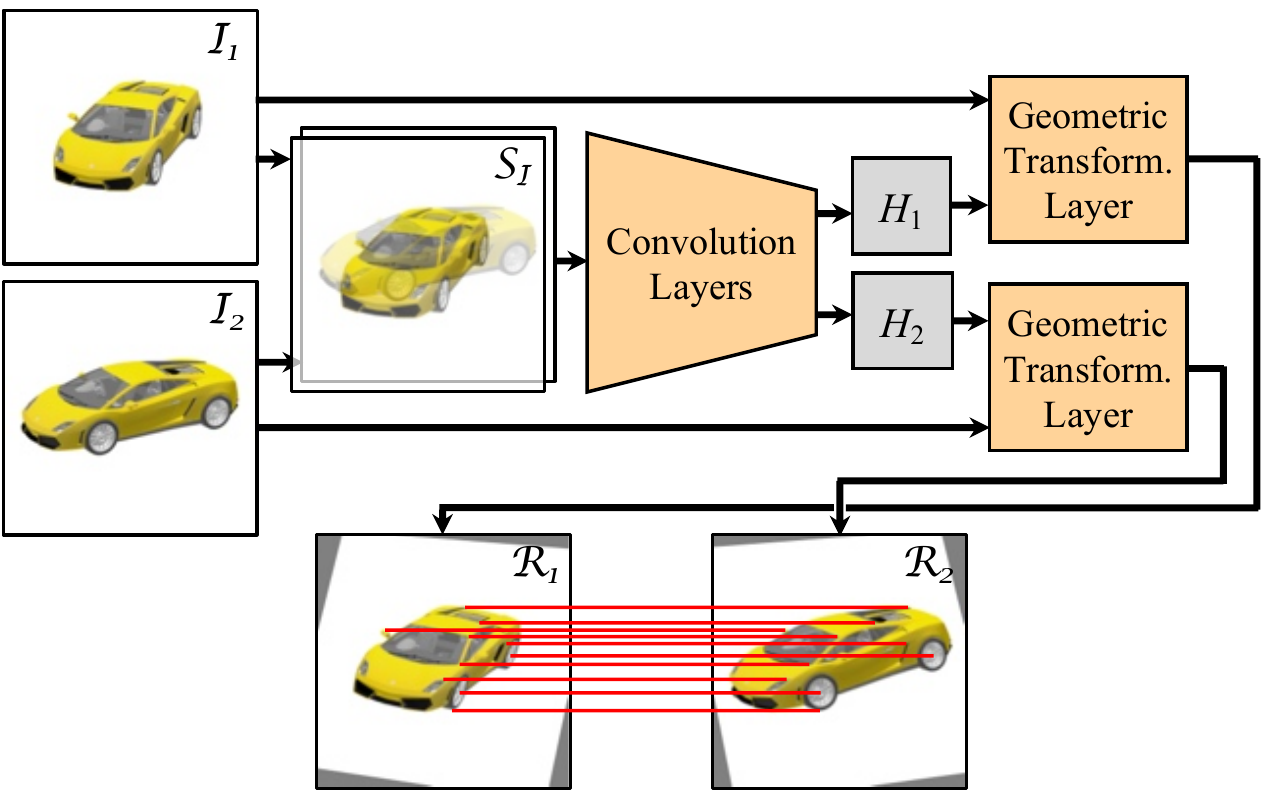}
\caption{Rectification network of Deep View Morphing. $\mathcal{I}_1$ and $\mathcal{I}_2$ are stacked to be 6-channel input $\mathcal{S}_I$. The last convolution layer outputs two homographies $H_1$ and $H_2$ to be applied to $\mathcal{I}_1$ and $\mathcal{I}_2$, respectively, via geometric transformation layers. The final output of the rectification network is a rectified pair $\mathcal{R}_1$ and $\mathcal{R}_2$. Red horizontal lines are shown to highlight several corresponding points between $\mathcal{R}_1$ and $\mathcal{R}_2$ that lie over horizontal epipolar lines.}
\label{fig:rectification_network}
\end{figure}

Figure~\ref{fig:rectification_network} shows the CNN architecture of the rectification network. We first stack two input images $\mathcal{I}_1$ and $\mathcal{I}_2$ to obtain 6-channel input $\mathcal{S}_{\mathcal{I}}$. Then convolution layers together with ReLU and max pooling layers process the stacked input $\mathcal{S}_{\mathcal{I}}$ to generate two homographies $H_1$ and $H_2$ in the form of 9D vectors. Finally, geometric transformation layers generate a rectified pair $\mathcal{R}_1$ and $\mathcal{R}_2$ by applying $H_1$ and $H_2$ to $\mathcal{I}_1$ and $\mathcal{I}_2$, respectively. The differentiation of the geometric transformation by homographies is straightforward and can be found in Appendix B. 

%%%%%%%%% Encoder-Decoder Network
\subsection{Encoder-decoder network}
\label{sec:encoder-decoder}

\begin{figure}
\centering
\includegraphics[width=3.2in]{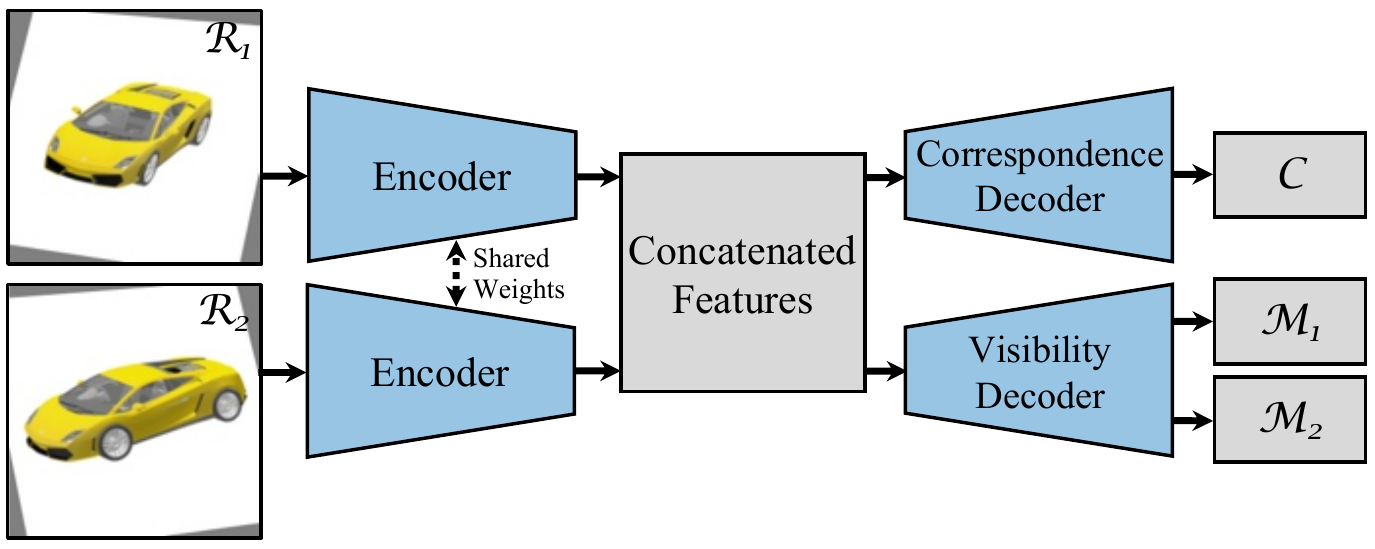}
\caption{Encoder-decoder network of Deep View Morphing. Each of two encoders sharing weights processes each of the rectified pair. The correspondence decoder and visibility decoder take in the concatenated encoder features and output the dense correspondences $\mathcal{C}$ and blending masks $\mathcal{M}_1$ and $\mathcal{M}_2$, respectively.}
\label{fig:encoder-decoder}
\end{figure} 

%%%%%%%%% Encoder
{\setlength{\parindent}{0cm}{\bf Encoders.}} The main role of encoders shown in Fig.~\ref{fig:encoder-decoder} is to encode correlations between two input images $\mathcal{R}_1$ and $\mathcal{R}_2$ into CNN features. There are two encoders sharing weights, each of which processes each of the rectified pair with convolution layers followed by ReLU and max pooling. The CNN features from the two encoders are concatenated channel-wise by the late fusion and fed into the correspondence decoder and visibility decoder.\\

%%%%%%%%% Correspondence Decoder
\begin{figure}
\centering
\includegraphics[width=2.4in]{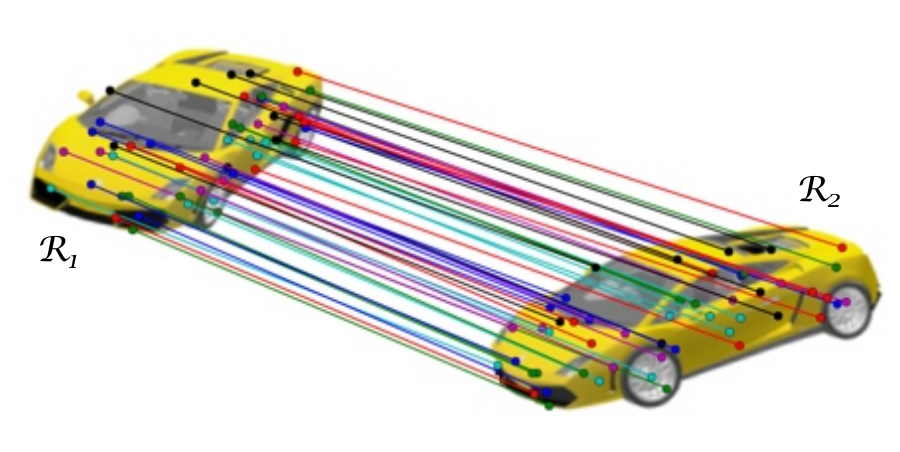}
\caption{Example of dense correspondences between $\mathcal{R}_1$ and $\mathcal{R}_2$ predicted by the correspondence decoder. For better visualization, $\mathcal{R}_2$ is placed lower than $\mathcal{R}_1$ and only 50 correspondences that are randomly chosen on the foreground are shown.}
\label{fig:correspondence}
\end{figure}

{\setlength{\parindent}{0cm}{\bf Correspondence decoder.}} The correspondence decoder shown in Fig.~\ref{fig:encoder-decoder} processes the concatenated encoder features by successive deconvolution layers as done in \cite{dosovitskiy2015flownet,Dosovitiskiy,Yang,tatarchenko2016multi,Zhou_eccv}. The last layer of the correspondence decoder is a convolution layer and outputs the dense correspondences $\mathcal{C}$ between $\mathcal{R}_1$ and $\mathcal{R}_2$. As $\mathcal{R}_1$ and $\mathcal{R}_2$ are already rectified by the rectification network, the predicted correspondences are only 1D, {\em i.e.}, correspondences along the same rows. 

Assume that $\mathcal{C}$ is defined with respect to the pixel coordinates $p$ of $\mathcal{R}_1$. We can then represent the point correspondence sets $P_1 = \{p_1^1, \ldots, p_1^M\}$ and $P_2 = \{p_2^1, \ldots, p_2^M\}$ as
\begin{equation}
p_1^i = p^i, \textrm{ }\textrm{ } p_2^i = p^i + \mathcal{C} (p^i), \textrm{ } i =1, \ldots, M,
\label{eqn:dense_corr}
\end{equation}
where $M$ is the number of pixels in $\mathcal{R}_1$. With these $P_1$ and $P_2$, we can now synthesize a middle view $\mathcal{R}_{\alpha=0.5}$ by (\ref{eqn:interpolation}). 

In (\ref{eqn:interpolation}), obtaining $\mathcal{R}_2 (p_2^i )$ needs interpolation because $p_2^i = p^i + \mathcal{C}(p^i)$ are generally non-integer valued. Such interpolation can be done very efficiently as it is sampling from regular grids. We also need to sample $\mathcal{R}_{\alpha=0.5}(q)$ on regular grid coordinates $q$ from $\mathcal{R}_{\alpha=0.5}(0.5 p_1^i + 0.5 p_2^i)$ as $0.5 p_1^i + 0.5 p_2^i$ are non-integer valued. Unlike $\mathcal{R}_2(p_2^i)$, sampling $\mathcal{R}_{\alpha=0.5}(q)$ from $\mathcal{R}_{\alpha=0.5}(0.5 p_1^i + 0.5 p_2^i)$ can be tricky because it is sampling from irregularly placed samples. 

To overcome this issue of sampling from irregularly placed samples, we can define $\mathcal{C}$ differently: $\mathcal{C}$ is defined with respect to the pixel coordinates $q$ of $\mathcal{R}_{\alpha=0.5}$. That is, the point correspondence sets $P_1$ and $P_2$ are obtained as
\begin{equation}
p_1^i = q^i + \mathcal{C}(q^i), \textrm{ }\textrm{ } p_2^i = q^i - \mathcal{C}(q^i), \textrm{ } i = 1, \ldots, M.
\label{eqn:our_dense_corr}
\end{equation}
Then the middle view $\mathcal{R}_{\alpha=0.5}$ can be easily synthesized as
\begin{equation}
\mathcal{R}_{\alpha=0.5}(q) = 0.5 \mathcal{R}_1(P_1) + 0.5 \mathcal{R}_2(P_2),
\label{eqn:our_interpolation}
\end{equation} 
where both $\mathcal{R}_1(P_1)$ and $\mathcal{R}_2(P_2)$ can be efficiently sampled.

Figure~\ref{fig:correspondence} shows an example of the dense correspondences between $\mathcal{R}_1$ and $\mathcal{R}_2$ predicted by the correspondence decoder. It is notable that the predicted correspondences are highly accurate even for textureless regions.\\

\begin{figure}
\centering
\includegraphics[width=3.3in]{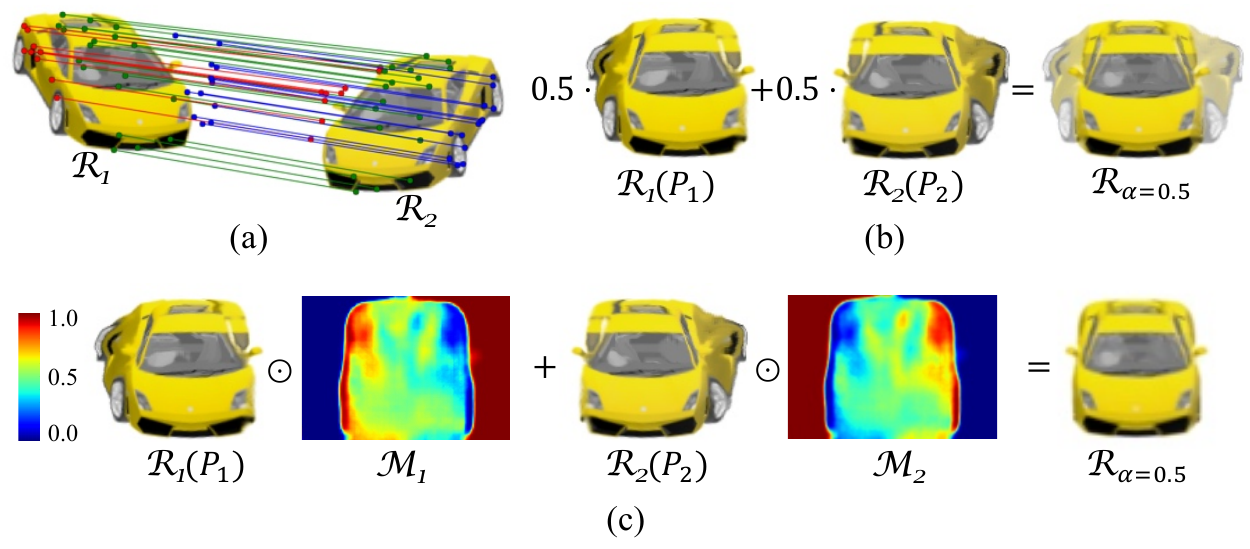}
\caption{(a) The correspondences for commonly visible regions are predicted accurately (green), but those for regions only visible in $\mathcal{R}_1$ or $\mathcal{R}_2$ are ill-defined and cannot be predicted correctly (red and blue). (b) The middle view synthesized by (\ref{eqn:our_interpolation}) using all of the correspondences suffers from severe ghosting artifacts. (c) The blending masks $\mathcal{M}_1$ and $\mathcal{M}_2$ generated by the visibility decoder correctly predict the visibility of pixels of $\mathcal{R}_1(P_1)$ and $\mathcal{R}_2(P_2)$ in the middle view, and thus we can obtain the ghosting-free middle view by (\ref{eqn:visibility}). For example, the left side of the car in $\mathcal{R}_1(P_1)$ has very low value in $\mathcal{M}_1$ close to 0 (dark blue) as it should not appear in the middle view while the corresponding region in $\mathcal{R}_2(P_2)$ is the background that should appear in the middle view and hence very high value in $\mathcal{M}_2$ close to 1 (dark red).}
\label{fig:visibility}
\end{figure}

%%%%%%%%%% Visibility Decoder
{\setlength{\parindent}{0cm}{\bf Visibility decoder.}} It is not unusual for $\mathcal{R}_1$ and $\mathcal{R}_2$ to have different visibility patterns as shown in Fig.~\ref{fig:visibility}(a). In such cases, the correspondences of pixels only visible in either one of views are ill-defined and thus cannot be predicted correctly. The undesirable consequence of using (\ref{eqn:our_interpolation}) with all of the correspondences for such cases is severe ghosting artifacts as shown in Fig.~\ref{fig:visibility}(b). 

In order to solve this issue, we adopt the idea to use blending masks proposed in~\cite{Zhou_eccv}. We use the visibility decoder shown in Fig.~\ref{fig:encoder-decoder} to predict visibility of each pixel of $\mathcal{R}_1(P_1)$ and $\mathcal{R}_2(P_2)$ in the synthesized view $\mathcal{R}_{\alpha=0.5}$. The visibility decoder processes the concatenated encoder features by successive deconvolution layers. At the end of the visibility decoder, a convolution layer outputs 1-channel feature map $\mathcal{M}$ that is converted to a blending mask $\mathcal{M}_1$ for $\mathcal{R}_1(P_1)$ by a sigmoid function. A blending mask $\mathcal{M}_2$ for $\mathcal{R}_2(P_2)$ is determined by $\mathcal{M}_2 = 1 - \mathcal{M}_1$. $\mathcal{M}_1$ and $\mathcal{M}_2$ represent the probability of each pixel of $\mathcal{R}_1(P_1)$ and $\mathcal{R}_2(P_2)$ to appear in the synthesized view $\mathcal{R}_{\alpha=0.5}$. 

Now we can synthesize the middle view $\mathcal{R}_{\alpha=0.5}$ using all of the correspondences and $\mathcal{M}_1$  and $\mathcal{M}_2$ as
\begin{equation}
\mathcal{R}_{\alpha=0.5}(q) = \mathcal{R}_1(P_1) \odot \mathcal{M}_1 + \mathcal{R}_2(P_2) \odot \mathcal{M}_2,
\label{eqn:visibility}
\end{equation}
where $\odot$ represents element-wise multiplication. As shown in Fig~\ref{fig:visibility}(c), regions that should not appear in the middle view have very low values close to 0 (dark blue) in $\mathcal{M}_1$ and $\mathcal{M}_2$ while commonly visible regions have similar values around 0.5 (green and yellow). As a result, we can obtain ghosting-free $\mathcal{R}_{\alpha=0.5}$ by (\ref{eqn:visibility}) as shown in Fig.~\ref{fig:visibility}(c). 

%%%%%%%%% View Morphing Network
\subsection{View morphing network}
\label{sec:view_morphing_network}

\begin{figure}
\centering
\includegraphics[width=3.15in]{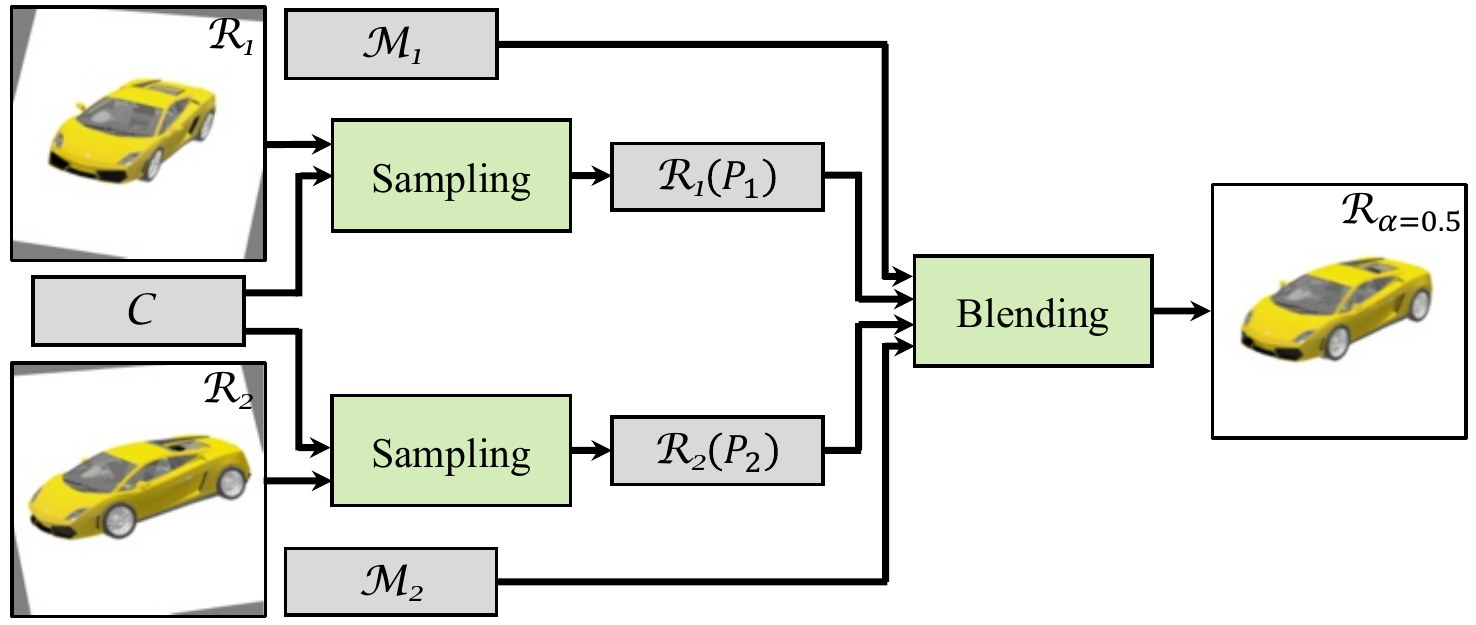}
\caption{View morphing network of Deep View Morphing. Sampling layers output $\mathcal{R}_1(P_1)$ and $\mathcal{R}_2(P_2)$ by sampling from $\mathcal{R}_1$ and $\mathcal{R}_2$ based on the dense correspondences $\mathcal{C}$. Then the blending layer synthesizes a middle view $\mathcal{R}_{\alpha=0.5}$ via (\ref{eqn:visibility}).}
\label{fig:view_morphing_network}
\end{figure}

Figure~\ref{fig:view_morphing_network} shows the view morphing network. Sampling layers take in the dense correspondences $\mathcal{C}$ and the rectified pair $\mathcal{R}_1$ and $\mathcal{R}_2$, and output $\mathcal{R}_1(P_1)$ and $\mathcal{R}_2(P_2)$ in (\ref{eqn:visibility}) by sampling pixel values of $\mathcal{R}_1$ and $\mathcal{R}_2$ at $P_1$ and $P_2$ determined by (\ref{eqn:our_dense_corr}). Here, we can use 1D interpolation for the sampling because  $\mathcal{C}$ represents 1D correspondences on the same rows. Then the blending layer synthesizes the middle view $\mathcal{R}_{\alpha=0.5}$ from $\mathcal{R}_1(P_1)$ and $\mathcal{R}_2(P_2)$ and their corresponding blending masks $\mathcal{M}_1$ and $\mathcal{M}_2$ by (\ref{eqn:visibility}). The view morphing network does not have learnable weights as both sampling and blending are fixed operations. 

%%%%%%%%% Network Training
\subsection{Network training}
All layers of DVM are differentiable and thus end-to-end-training with a single loss at the end comparing the synthesized middle view and ground truth middle view is possible. For training, we use the Euclidean loss defined as 
\begin{equation}
L = \sum_{i=1}^{M} \frac{1}{2} || \mathcal{R}_{\alpha=0.5}(q^i) - \mathcal{R}_{\textrm{GT}}(q^i) ||_2^2,
\label{eqn:loss}
\end{equation}
where $\mathcal{R}_{\textrm{GT}}$ is the desired ground truth middle view image and $M$ is the number of pixels. Note that we do not need the post-warping step as in \cite{Seitz_1996_siggraph} (Section~\ref{sec:post-warping}) because the rectification network is trained to rectify $\mathcal{I}_1$ and $\mathcal{I}_2$ so that the middle view of $\mathcal{R}_1$ and $\mathcal{R}_2$ can be directly matched against the desired ground truth middle view $\mathcal{R}_{\textrm{GT}}$. 

%%%%%%%%% Implementation Details
\subsection{Implementation details}
The CNN architecture details of DVM such as number of layers and kernel sizes and other implementation details are shown in Appendix A. With Intel Xeon E5-2630 and a single Nvidia Titan X, DVM processes a batch of 20 input pairs of $224 \times 224$ in 0.269 secs using the modified version of Caffe \cite{jia2014caffe}. 
\section{Experiments}
\label{sec:exp}

We now demonstrate the view synthesis performance of DVM via experiments using two datasets: (i) ShapeNet \cite{shapenet} and (ii) Multi-PIE \cite{MultiPIE}. We mainly compare the performance of DVM with that of ``View Synthesis by Appearance Flow" (VSAF) \cite{Zhou_eccv}. We evaluated VSAF using the codes kindly provided by the authors. For training of both methods, we initialized all weights by the Xavier method \cite{Xavier} and all biases by constants of $0.01$, and used the Adam solver \cite{Adam} with $\beta_1=0.9$ and $\beta_2=0.999$ with the mini-batch sizes of 160 and initial learning rates of $0.0001$. 

\subsection{Experiment 1: ShapeNet}
\label{sec:experiment1_shapenet}

{\setlength{\parindent}{0cm}{\bf Training data.}} We used ``Car", ``Chair", ``Airplane", and ``Vessel" of ShapeNet to create training data. We randomly split all 3D models of each category into $80\%$ training and $20\%$ test instances. We rendered each model using Blender (https://www.blender.org) using cameras at azimuths of $0^{\circ}$ to $355^{\circ}$ with $5^\circ$ steps and elevations of $ 0^{\circ}$ to $30^{\circ}$ with $10^{\circ}$ steps with the fixed distance to objects. We finally cropped object regions using the same central squares for all viewpoints and resized them to $224\times224$. 

We created training triplets $\{ \mathcal{I}_1,  \mathcal{R}_{\textrm{GT}}, \mathcal{I}_2 \}$ where $\mathcal{I}_1$, $\mathcal{R}_{\textrm{GT}}$, and $\mathcal{I}_2$ have the same elevations. Let $\phi_1$, $\phi_2$, and $\phi_{\textrm{GT}}$ denote azimuths of $\mathcal{I}_1$, $\mathcal{I}_2$, and $\mathcal{R}_{\textrm{GT}}$. We first sampled $\mathcal{I}_1$ with $\phi_1$ multiples of $10^\circ$, and sampled $\mathcal{I}_2$ to satisfy $\Delta \phi = \phi_2 - \phi_1 = \{20^{\circ}, 30^{\circ}, 40^{\circ}, 50^{\circ}\}$. $\mathcal{R}_{\textrm{GT}}$ is then selected to satisfy $\phi_{\textrm{GT}} - \phi_1 = \phi_2 - \phi_{\textrm{GT}} = \{10^{\circ}, 15^{\circ}, 20^{\circ}, 25^{\circ}\}$. We provided VSAF with 8D one-hot vectors \cite{Zhou_eccv} to represent viewpoint transformations from $\mathcal{I}_1$ to $\mathcal{R}_{\textrm{GT}}$ and from $\mathcal{I}_2$ to $\mathcal{R}_{\textrm{GT}}$ equivalent to azimuth differences of $\{ \pm10^{\circ}, \pm 15^{\circ}, \pm 20^{\circ}, \pm 25^{\circ} \}$. The number of training triplets for ``Car", ``Chair", ``Airplane", and ``Vessel" are about 3.4M, 3.1M, 1.9M, and 0.9M, respectively. More details of the ShapeNet training data are shown in Appendix C.\\

{\setlength{\parindent}{0cm}{\bf Category-specific training.}} We first show view synthesis results of DVM and VSAF trained on each category separately. Both DVM and VSAF were trained using exactly the same training data. For evaluating the view synthesis results, we randomly sampled 200,000 test triplets for each category created with the same configuration as that of the training triplets. As an error metric, we use the mean squared error (MSE) between the synthesized output and ground truth summed over all pixels.

\begin{table}[]
\centering
\caption{Mean of MSE by DVM and VSAF trained for ``Car", ``Chair", ``Airplane", and ``Vessel" in a category-specific way.} 
\label{tab:MSE_shapenet}
\small
\begin{tabular}{ccccc}
\\
\hline
 & Car & Chair & Airplane & Vessel\\
\hline
DVM  & {\bf 44.70} & {\bf 61.00} & {\bf 22.30} & {\bf 42.74}\\
VSAF & 70.11 & 140.35 & 46.80 & 95.99\\
\hline
\end{tabular}
\end{table}

%\begin{table}[]
%\centering
%\caption{Mean and standard deviation of MSE by DVM and VSAF trained for ``Car", ``Chair", ``Airplane", and ``Vessel" in a category-specific way.} 
%\label{tab:MSE_shapenet}
%\small
%\begin{tabular}{ccc}
%\\
%\hline
% & DVM & VSAF\\
%\hline
%Car &{\bf 44.70 ($\pm$42.69)} & 70.11 ($\pm$68.10)\\
%Chair &{\bf 61.00 ($\pm$82.71)} & 140.35 ($\pm$174.86)\\
%Airplane & {\bf 22.30 ($\pm$32.38)} & 46.80 ($\pm$68.55)\\
%Vessel & {\bf 42.74 ($\pm$77.75)} & 95.99 ($\pm$147.29)\\
%\hline
%\end{tabular}
%\end{table}

\begin{figure}
\centering
\includegraphics[width=3.2in]{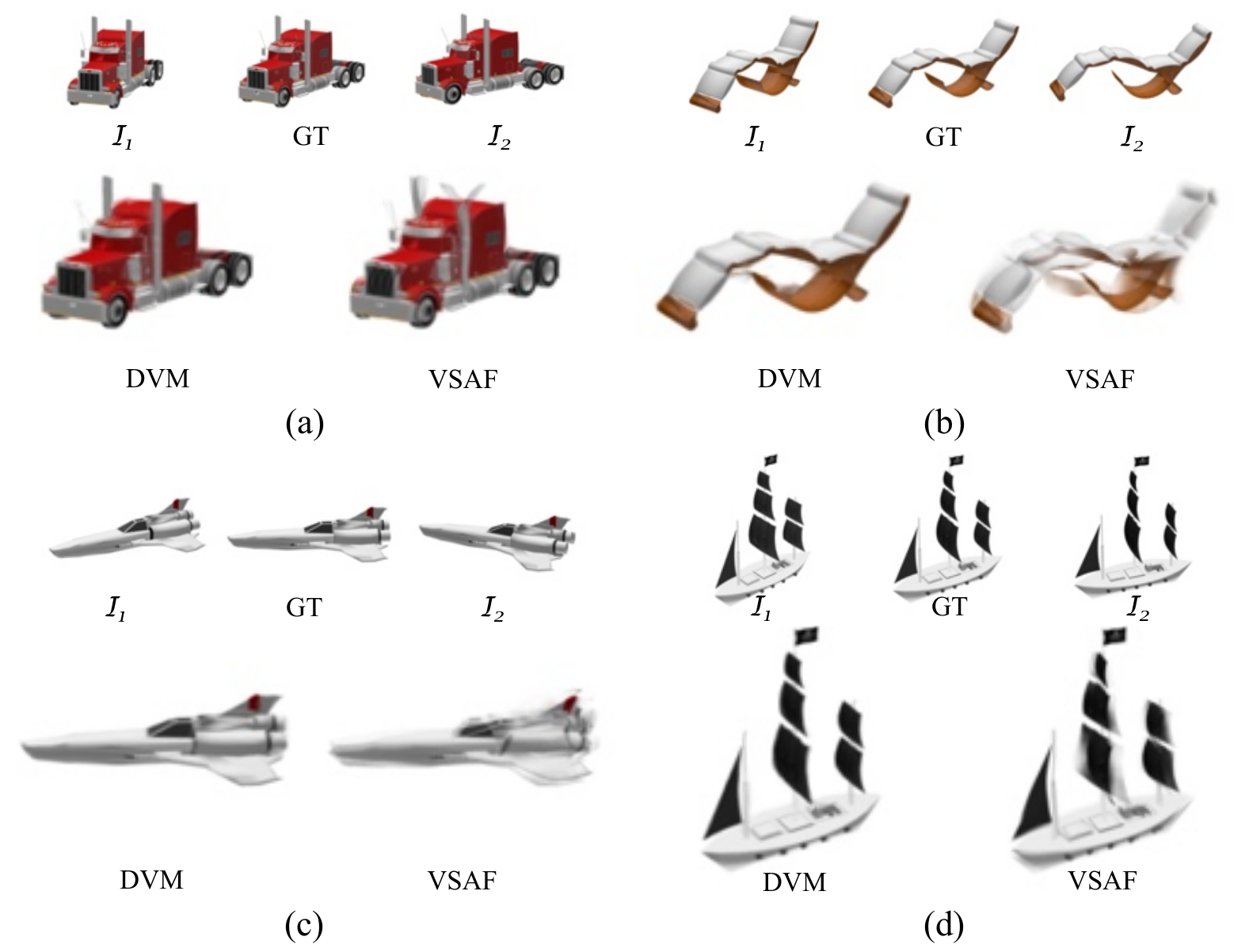}
\caption{Comparisons of view synthesis results by DVM and VSAF on test samples of (a) ``Car", (b) ``Chair", (c) ``Airplane", and (d) ``Vessel" of ShapeNet. Two input images are shown on the left and right sides of the ground truth image (``GT"). More comparisons are shown in Appendix C.}
\label{fig:results_shapenet}
\end{figure}

\begin{figure}
\centering
\includegraphics[width=3.2in]{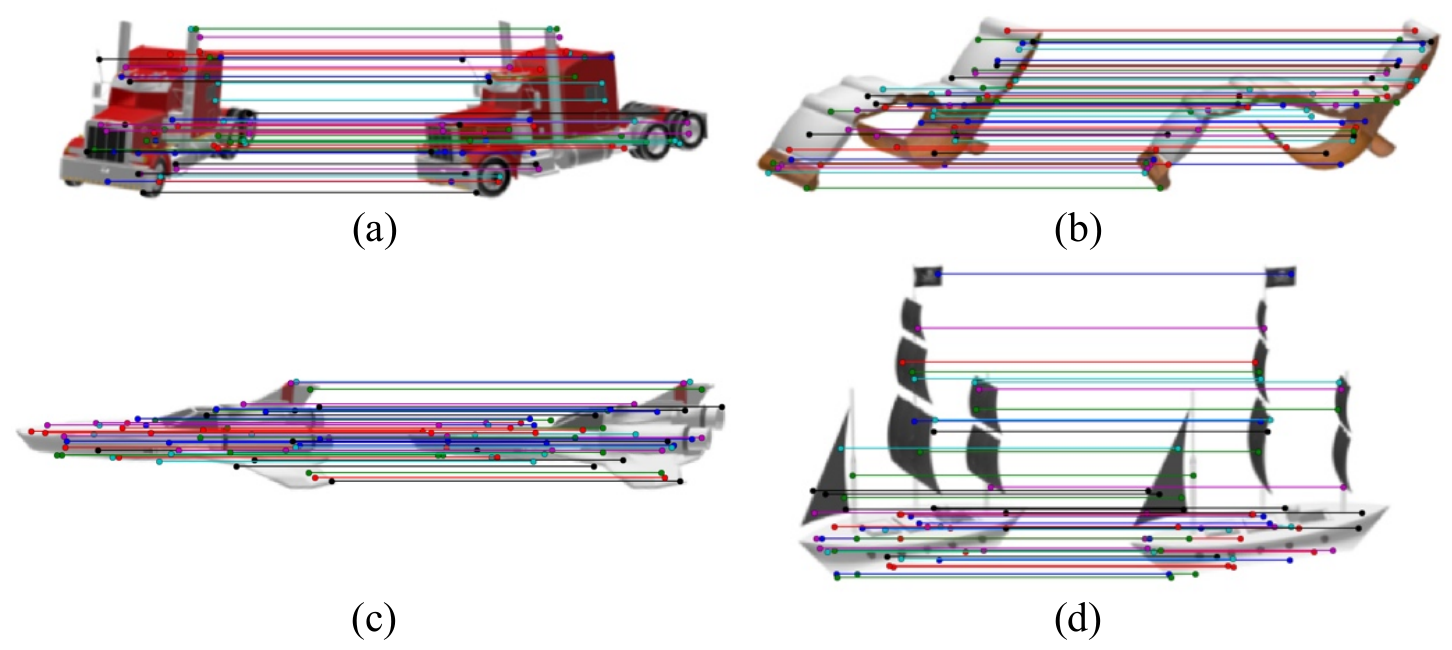}
\caption{Examples of rectification results and dense correspondences obtained by DVM on the test input images shown in Fig.~\ref{fig:results_shapenet}. More examples are shown in Appendix C.}
\label{fig:corr_shapenet}
\end{figure}

\begin{figure}
\centering
\includegraphics[width=2.8in]{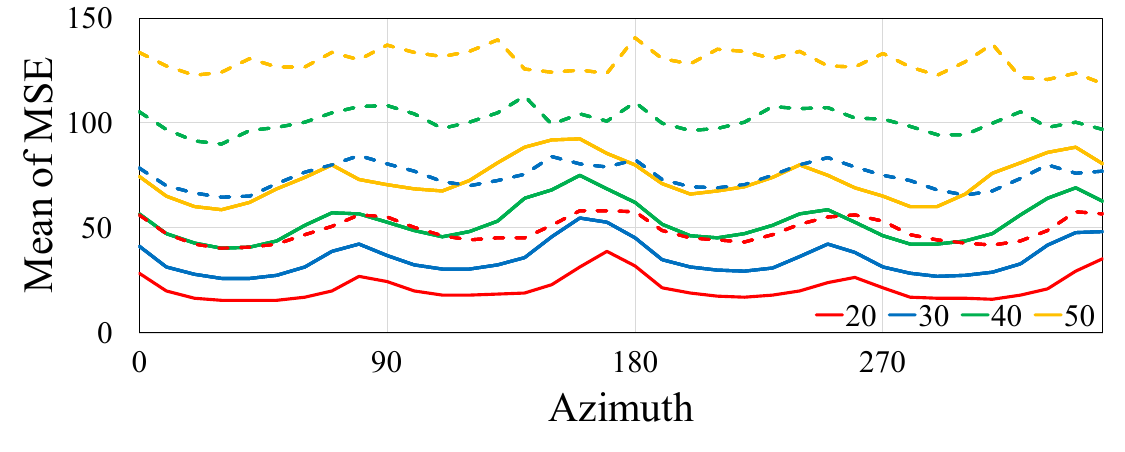}
\caption{Plots of mean of MSE by DVM (solid) and VSAF (dashed) as a function of $\phi_1$ (azimuth of $\mathcal{I}_1$) for all test triplets of ``Car", ``Chair", ``Airplane", and ``Vessel". Different line colors represent different azimuth differences $\Delta \phi$  between $\mathcal{I}_1$ and $\mathcal{I}_2$.}
\label{fig:errors_shapenet}
\end{figure}

Figure~\ref{fig:results_shapenet} shows qualitative comparisons of view synthesis results by DVM and VSAF. It is clear that view synthesis results by DVM are visually more pleasing with much less ghosting artifacts and much closer to the ground truth views than those by VSAF. Table~\ref{tab:MSE_shapenet} shows the mean of MSE by DVM and VSAF for each category. The mean of MSE by DVM are considerably smaller than those by VSAF for all four categories, which matches well the qualitative comparisons in Fig.~\ref{fig:results_shapenet} . The mean of MSE by DVM for ``Car", ``Chair", ``Airplane", and ``Vessel" are $63.8\%$,  $43.5\%$,  $47.6\%$, and $44.5\%$ of that by VSAF, respectively.

Figure~\ref{fig:corr_shapenet} shows the rectification results and dense correspondences obtained by DVM for the test input images shown in Fig.~\ref{fig:results_shapenet}. Note that DVM yields highly accurate rectification results and dense correspondence results. In fact, it is not possible to synthesize the middle view accurately if one of them is incorrect. The quantitative analysis of the rectification accuracy by DVM is shown in Appendix C.

Figure~\ref{fig:errors_shapenet} shows plots of mean of  MSE by DVM and VSAF as a function of $\phi_1$ (azimuth of $\mathcal{I}_1$) where different line colors represent different azimuth differences $\Delta \phi$ between $\mathcal{I}_1$ and $\mathcal{I}_2$. As expected, the mean of MSE increases as $\Delta \phi$ increases. Note that the mean of MSE by DVM for $\Delta \phi = 50^\circ$ is similar to that by VSAF for $\Delta \phi = 30^\circ$. Also note that the mean of MSE by DVM for each $\Delta \phi$ has peaks near $\phi_1 = 90^\circ \cdot i - \Delta \phi /2, i=0,1,2,3$, where there is considerable visibility changes between $\mathcal{I}_1$ and $\mathcal{I}_2$, {\em e.g.}, from a right-front view $\mathcal{I}_1$ to a left-front view $\mathcal{I}_2$.

We also compare the performance of DVM and VSAF trained for the larger azimuth differences up to $90^\circ$. Due to the limited space, the results are shown in Appendix C.\\

{\setlength{\parindent}{0cm}{\bf Robustness test.}} We now test the robustness of DVM and VSAF to inputs that have different azimuths and elevations from those of the training data. We newly created 200,000 test triplets of ``Car" with azimuths and elevations that are $5^\circ$ shifted from those of the training triplets but still with $\Delta \phi = \{20^\circ, 30^\circ, 40^\circ, 50^\circ\}$. The mean of MSE for the  $5^\circ$ shifted test triplets by DVM and VSAF are 71.75 and 107.64, respectively. Compared to the mean of MSE by DVM and VSAF on the original test triplets of ``Car" in Tab.~\ref{tab:MSE_shapenet}, both DVM and VSAF performed worse similarly: $61\%$ MSE increase by DVM and $54\%$ MSE increase by VSAF. However, note that the mean of MSE by DVM on the $5^\circ$ shifted test triplets (71.75) is similar to that by VSAF on the original test triplets (70.11). 

\begin{figure}
\centering
\includegraphics[width=2.8in]{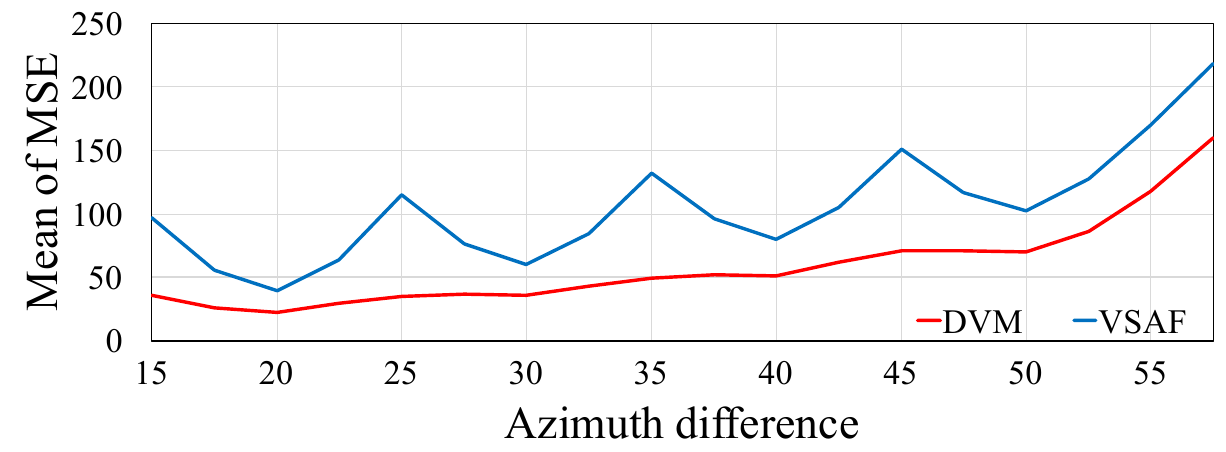}
\caption{Plots of mean of MSE by DVM (red) and VSAF (blue) as a function of azimuth difference $\Delta \phi$  between $\mathcal{I}_1$ and $\mathcal{I}_2$ for ``Car". Here, the azimuth differences are $15^\circ \leq \Delta \phi < 60^\circ$ with $2.5^\circ$ steps.}
\label{fig:errors_shapenet_additional_vp}
\end{figure}

We also test the robustness of DVM and VSAF to inputs with azimuth differences $\Delta \phi$ that are different from those of the training data. We newly created 500,000 test triplets of ``Car" with $\mathcal{I}_1$ that are the same as the training triplets and $\mathcal{I}_2$ and $\mathcal{R}_{\textrm{GT}}$ corresponding to $15^\circ \leq \Delta \phi < 60^\circ$ with $2.5^\circ$ steps. We provided VSAF with 8D one-hot vectors by finding the elements from $\{ \pm10^{\circ}, \pm 15^{\circ}, \pm 20^{\circ}, \pm 25^{\circ} \}$ closest to $\phi_1 - \phi_{\textrm{GT}}$ and $\phi_2 - \phi_{\textrm{GT}}$.

Figure~\ref{fig:errors_shapenet_additional_vp} shows plots of mean of MSE by DVM and VSAF for the new 500,000 test triplets of ``Car". It is clear that DVM is much more robust to the unseen $\Delta \phi$ than VSAF. VSAF yielded much higher MSE for the unseen $\Delta \phi$ compared to that for $\Delta \phi$ multiples of 10. Contrarily, the MSE increase by DVM for such unseen $\Delta \phi$ is minimal except for $\Delta \phi > 50^\circ$. This result suggests that DVM which directly considers two input images together for synthesis without relying on the viewpoint transformation inputs has more generalizability than VSAF.\\ 

{\setlength{\parindent}{0cm}{\bf Category-agnostic training.}} We now show view synthesis results of DVM and VSAF trained in a category-agnostic way, {\em i.e.}, we trained DVM and VSAF using all training triplets of all four categories altogether. For this category-agnostic training, we limited the maximum number of training triplets of each category to 1M. For testing, we additionally selected four unseen categories from ShapeNet: ``Motorcycle", ``Laptop", ``Clock", and ``Bookshelf". The test triplets of the unseen categories were created with the same configuration as that of the training triplets.

\begin{figure}
\centering
\includegraphics[width=3.2in]{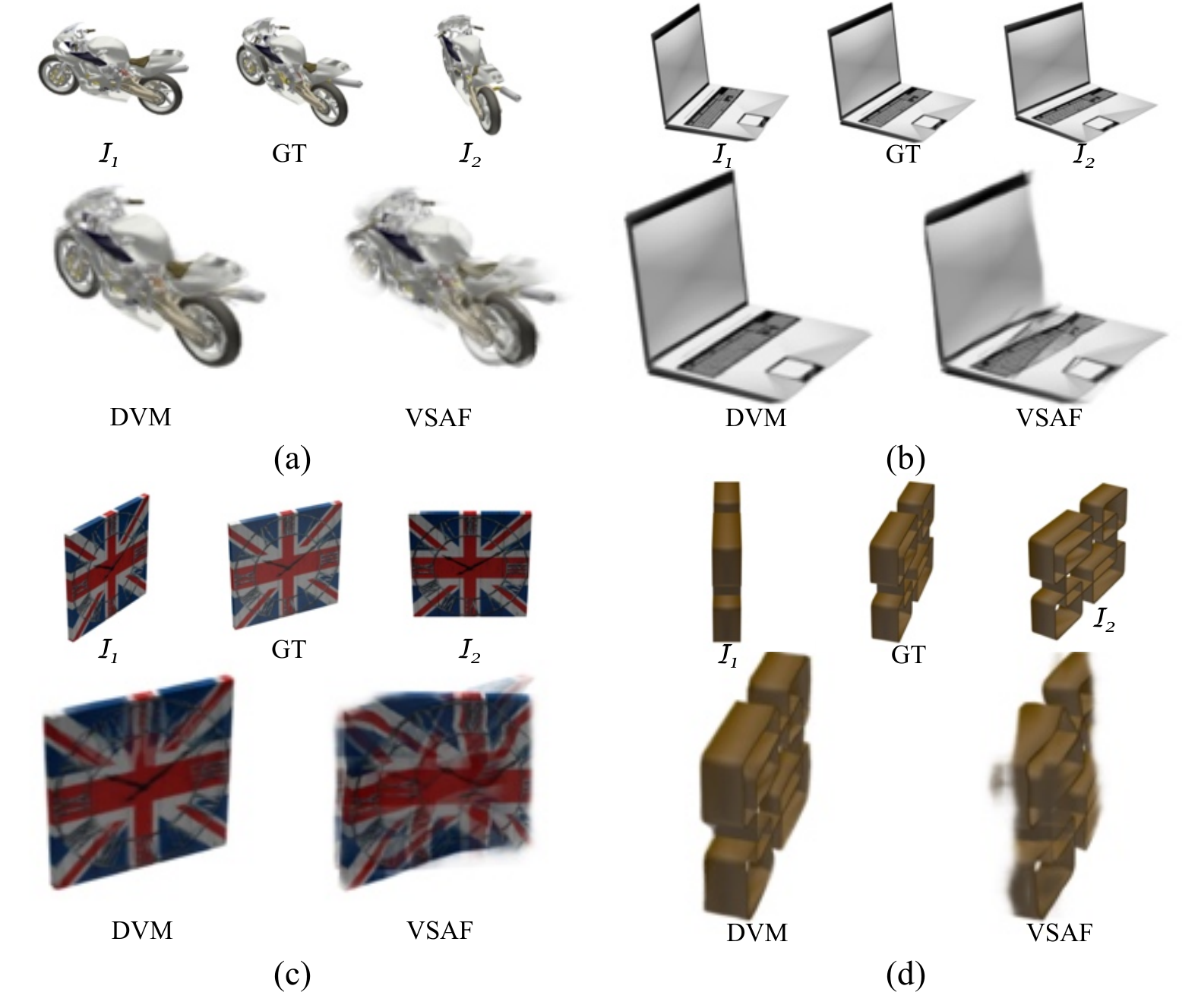}
\caption{Comparisons of view synthesis results by DVM and VSAF on test samples of unseen (a) ``Motorcycle", (b) ``Laptop", (c) ``Clock", and (d) ``Bookshelf" of ShapeNet. More comparisons are shown in Appendix C.}
\label{fig:results_shapenet_unseen}
\end{figure}

\begin{table}[]
\centering
\caption{Mean of MSE by DVM and VSAF trained for ``Car", ``Chair", ``Airplane", and ``Vessel" in a category-agnostic way.} 
\label{tab:MSE_shapenet_all}
\small
\begin{tabular}{ccccc}
\\
\hline
 & Car & Chair & Airplane & Vessel\\
\hline
DVM  & {\bf 52.56} & {\bf 73.01} & {\bf 24.73} & {\bf 38.42}\\
VSAF & 83.36 & 161.59 & 51.95 & 88.47\\
\hline
\hline
 & Motorcycle & Laptop & Clock & Bookshelf\\
\hline
DVM  & {\bf 154.45} & {\bf 102.27} & {\bf 214.02} & {\bf 171.81}\\
VSAF & 469.01 & 262.33 & 491.82 & 520.22\\
\hline
\end{tabular}
\end{table}

%\begin{table}[]
%\centering
%\caption{Mean and standard deviation of MSE by DVM and VSAF trained for ``Car", ``Chair", ``Airplane", and ``Vessel" in a category-agnostic way.} 
%\label{tab:MSE_shapenet_all}
%\small
%\begin{tabular}{ccc}
%\\
%\hline
% & DVM & VSAF \\
%\hline
%Car & {\bf 52.56 ($\pm$47.28)} & 83.36 ($\pm$75.07)\\
%Chair & {\bf 73.01 ($\pm$92.96)} & 161.59 ($\pm$182.65)  \\
%Airplane & {\bf 24.73 ($\pm$31.66)}  &  51.95 ($\pm$66.91)\\
%Vessel & {\bf 38.42 ($\pm$67.76)}  & 88.47 ($\pm$130.41)\\
%Motorcycle& {\bf 154.45 ($\pm$120.94)}   & 469.01 ($\pm$306.14)\\
%Laptop & {\bf 102.27 ($\pm$163.72)} & 262.33 ($\pm$258.58)\\
%Clock & {\bf 214.02 ($\pm$458.56)}  & 491.82 ($\pm$602.02)\\
%Bookshelf & {\bf 171.81 ($\pm$270.55)}  & 520.22 ($\pm$567.45)\\
%\hline
%\end{tabular}
%\end{table}

Figure~\ref{fig:results_shapenet_unseen} shows qualitative comparisons of view synthesis results by DVM and VSAF on the unseen categories. We can see the view synthesis results by DVM are still highly accurate even for the unseen categories. Especially, DVM even can predict the blending masks correctly as shown in Fig.~\ref{fig:results_shapenet_unseen}(d). Contrarily, VSAF yielded view synthesis results with lots of ghosting artifacts and severe shape distortions.

Table~\ref{tab:MSE_shapenet_all} shows the mean of MSE by DVM and VSAF trained in a category-agnostic way. Compared to Tab.~\ref{tab:MSE_shapenet},  we can see the mean of MSE by both DVM and VSAF for ``Car", ``Chair", and ``Airplane" slightly increased due to less training samples of the corresponding categories. Contrarily, the mean of MSE by both DVM and VSAF for ``Vessel" decreased mainly due to the training samples of the other categories. The performance difference between DVM and VSAF for the unseen categories is much greater than that for the seen categories. The mean of MSE by DVM for ``Motorcycle", ``Laptop", ``Clock", and ``Bookshelf" are $32.9\%$,  $39.0\%$,  $43.5\%$, and $33.0\%$ of that by VSAF, respectively. These promising results by DVM on the unseen categories suggest that DVM can learn general features necessary for rectifying image pairs and establishing correspondences between them. The quantitative analysis of the rectification accuracy by DVM for the unseen categories is shown in Appendix C. 

\begin{figure}
\centering
\includegraphics[width=3in]{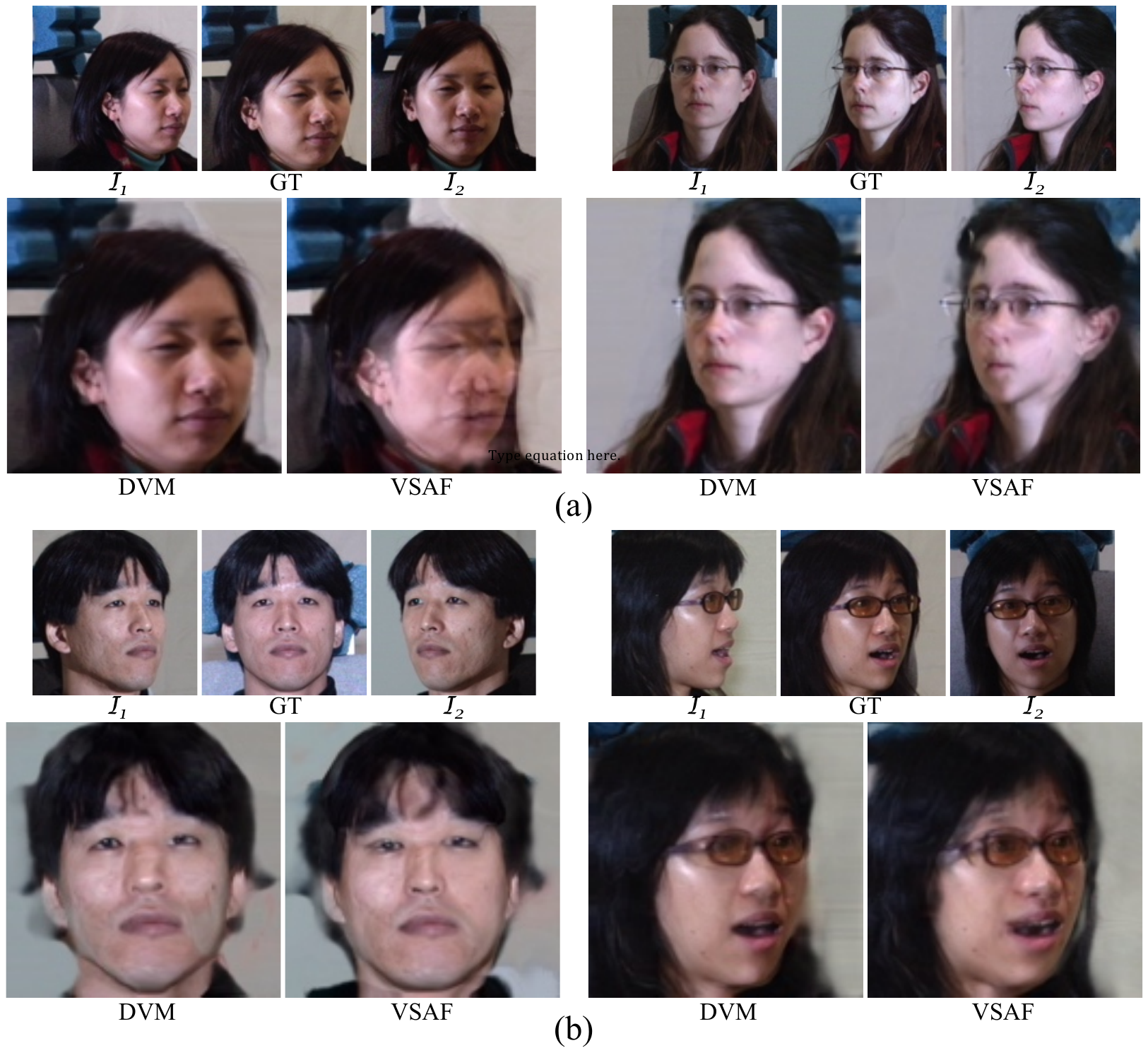}
\caption{Comparisons of view synthesis results by DVM and VSAF on Multi-PIE test samples with (a) loose and (b) tight crops. More comparisons are shown in Appendix C.}
\label{fig:results_face}
\end{figure}

\begin{figure*}
\centering
\includegraphics[width=6.2in]{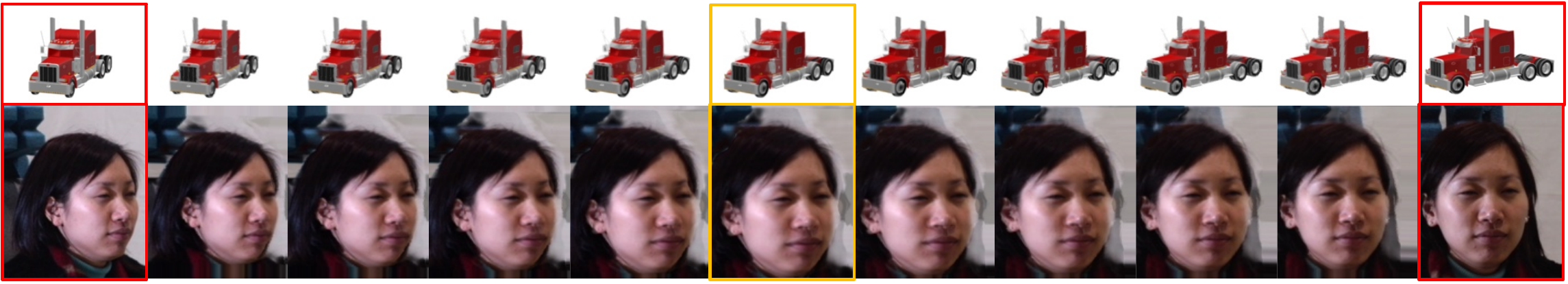}
\caption{Intermediate view synthesis results on the ShapeNet test input images shown in Fig.~\ref{fig:results_shapenet} (top) and the Multi-PIE test input images shown in Fig.~\ref{fig:results_face}(a) (bottom). Red and orange boxes represent input image pairs and $\mathcal{R}_{\alpha=0.5}$ directly generated by DVM, respectively. More intermediate view synthesis results are shown in Appendix C.}
\label{fig:interpolation_results}
\end{figure*}

\subsection{Experiment 2: Multi-PIE}
{\setlength{\parindent}{0cm}{\bf Training data.}} Multi-PIE dataset \cite{MultiPIE} contains face images of 337 subjects captured at 13 viewpoints from $0^{\circ}$ to $180^{\circ}$ azimuth angles. We split 337 subjects into 270 training and 67 test subjects. We used 11 viewpoints from $15^{\circ}$ to $165^{\circ}$ because images at $0^{\circ}$ and $180^{\circ}$ have drastically different color characteristics. We sampled $\mathcal{I}_1$ and $\mathcal{I}_2$ to have $\Delta \phi  = \{30^{\circ}, 60^{\circ}\}$, and picked $\mathcal{R}_{\textrm{GT}}$ to satisfy $\phi_{\textrm{GT}} - \phi_1 = \phi_2 - \phi_{\textrm{GT}} = \{15^{\circ}, 30^{\circ}\}$. The number of training triplets constructed in this way is 643,760. We provided VSAF with 4D one-hot vectors accordingly.

%\begin{figure}
%\centering
%\includegraphics[width=3.3in]{figs/loose_tight.pdf}
%\caption{Examples of Multi-PIE training triplets $\{\mathcal{I}_1, \mathcal{R}_{\textrm{GT}}, \mathcal{I}_2 \}$ with (a) loose and (b) tight facial region crops.}
%\label{fig:loose_tight}
%\end{figure}

Multi-PIE provides detailed facial landmarks annotations but only for subsets of whole images. Using those annotations, we created two sets of training data with (i) loose and (ii) tight facial region crops. For the loose crops, we used a single bounding box for all images of the same viewpoint that encloses all facial landmarks of those images. For the tight crops, we first performed face segmentation using FCN \cite{FCN} trained with convex hull masks of facial landmarks. We then used a bounding box of the segmented region for each image. For both cases, we extended bounding boxes to be square and include all facial regions and finally resized them to $224\times224$.\\%Figure~\ref{fig:loose_tight} shows examples of the training triplets with the loose and tight crops. \\

{\setlength{\parindent}{0cm}{\bf Results.}} We trained DVM and VSAF using each of the two training sets separately. For testing, we created two sets of 157,120 test triplets from 67 test subjects, one with the loose crops and the other with the tight crops, with the same configuration as that of the training sets. 

Figure~\ref{fig:results_face}(a) shows qualitative comparisons of view synthesis results by DVM and VSAF on the test triplets with the loose facial region crops. The view synthesis results by VSAF suffer lots of ghosting artifacts and severe shape distortions mainly because (i) faces are not aligned well and (ii) their scales can be different. Contrarily, DVM yielded very satisfactory view synthesis results by successfully dealing with the unaligned faces and scale differences thanks to the presence of the rectification network. These successful view synthesis results by DVM have significance in that DVM can synthesize novel views quite well even with the  the camera setup not as precise as that of the ShapeNet rendering and objects with different scales.

\begin{table}[]
\centering
\caption{Mean of MSE by DVM and VSAF for Multi-PIE test triplets with the loose and tight facial region crops.} 
\label{tab:MSE_multipie}
\small
\begin{tabular}{ccc}
\\
\hline
 & Loose facial region crops & Tight facial region crops\\
\hline
DVM  & {\bf 162.62} & {\bf 164.77} \\
VSAF & 267.83 & 194.30\\
\hline
\end{tabular}
\end{table}

%\begin{table}[]
%\centering
%\caption{Mean and standard deviation of MSE by DVM and VSAF for test triplets of Multi-PIE.} 
%\label{tab:MSE_multipie}
%\small
%\begin{tabular}{ccc}
%\\
%\hline
% & DVM & VSAF \\
%\hline
%Loose crops & {\bf 162.62 ($\pm$132.29)} &267.83 ($\pm$203.39) \\
%Tight crops & {\bf 164.77 ($\pm$129.44)} & 194.30 ($\pm$143.81)\\
%\hline
%\end{tabular}
%\end{table}

Figure~\ref{fig:results_face}(b) shows qualitative comparisons of view synthesis results by DVM and VSAF on the test triplets with the tight facial region crops. The view synthesis results by VSAF are much improved compared to the case of the loose facial region crops because the facial regions are aligned fairly well and their scale differences are negligible. However, the view synthesis results by DVM are still better than those by VSAF with less ghosting artifacts and less shape distortions. Table~\ref{tab:MSE_multipie} shows the mean of MSE by DVM and VSAF for the Multi-PIE test triplets that match well the qualitative comparisons in Fig.~\ref{fig:results_face}.

\subsection{Experiment 3: Intermediate view synthesis}
\label{sec:intermediate_view}

Figure~\ref{fig:interpolation_results} shows the intermediate view synthesis results obtained by linearly interpolating the blending masks $\mathcal{M}_1$ and $\mathcal{M}_2$ as well as $\mathcal{R}_1$ and $\mathcal{R}_2$.  As the dense correspondences predicted by DVM are highly accurate, we can synthesize highly realistic intermediate views. The detailed procedure to synthesize the intermediate views and more results are shown in Appendix C.

\section{Conclusion and discussion}
In this paper, we proposed DVM, a CNN-based view synthesis method inspired by View Morphing \cite{Seitz_1996_siggraph}. Two input images are first automatically rectified by the rectification network. The encoder-decoder network then outputs the dense correspondences between the rectified images and blending masks to predict the visibility of pixels of the rectified images in the middle view. Finally, the view morphing network synthesizes the middle view using the dense correspondences and blending masks. We experimentally showed that DVM can synthesize novel views with detailed textures and well-preserved geometric shapes clearly better than those by the CNN-based state-of-the-art.

Deep View Morphing still can be improved in some aspects. For example, it is generally difficult for Deep View Morphing to deal with very complex thin structures. Plus, the current blending masks cannot properly deal with the different illumination and color characteristics between input images, and thus blending seams can be visible in some cases. Examples of these challenging cases for DVM are shown in Appendix C. Future work will be focused on improving the performance for such cases. 

\section*{Acknowledgment}
We would like to thank Tinghui Zhou for kindly sharing codes of ``View Synthesis by Appearance Flow" \cite{Zhou_eccv} and helpful comments with us.  

{\small
\bibliographystyle{ieee}
\bibliography{cvpr17_bib}
}

%%%%%%%%% INTRODUCTION
\section*{Appendix A. CNN architecture details}
We present the CNN architecture details of each sub-network of Deep View Morphing. Note that only layers with learnable weights are shown here. As aforementioned in Section 3 of the main paper, we can consider the early and late fusions in the rectification network and encoder-decoder network. We define the early fusion as channel-wise concatenation of two input images while the late fusion as channel-wise concatenation of CNN features of the two input images processed by convolution layers. 

\subsection*{A.1. Rectification network}

\begin{table*}[]
\centering
\caption{CNN architecture details of the rectification network of Deep View Morphing. All convolution layers are followed by ReLU except for ``RC8", ``RC8-1" and ``RC8-2" whose output is the homographies used for rectification. $k$: kernel size ($k \times k$). $s$: stride in both horizontal and vertical directions. $c$: number of output channels. $d$: output spatial dimension ($d \times d$). Conv: convolution. MPool: max pooling. APool: average pooling. Concat: channel-wise concatenation.}
\label{tab:cnn_details}
\footnotesize
\begin{tabular}{cccccc|ccccccccccccccccc}
\\
\hline
\multicolumn{6}{c|}{{\bf Rectification network with early fusion}} &  \multicolumn{12}{c}{{\bf Rectification network with late fusion}}\\
\hline
Name & Type & $k$  & $s$ & $c$ & $d$ &Name &Type & $k$ &$s$ & $c$ & $d$ & Name & Type & $k$ &$s$ & $c$ & $d$ &\\
\hline
$\mathcal{S}_I$ & Input & $\cdot$ & $\cdot$ & 6 & 224 & $\mathcal{I}_1$ & Input & $\cdot$ & $\cdot$ &3 & 224& $\mathcal{I}_2$ & Input & $\cdot$ & $\cdot$ &3 & 224\\
RC1 & Conv & 9 & 2 & 32 & 112 &  RC1-1 & Conv& 9 & 2 & 32 & 112 & RC1-2 & Conv& 9 & 2 & 32 & 112\\
RP1 & MPool & 3 & 2 & 32 & 56 &  RP1-1& MPool& 3 & 2 & 32 & 56 & RP1-2& MPool& 3 & 2 & 32 & 56\\
RC2 & Conv &7 & 1 & 64 & 56 & RC2-1 & Conv & 7 & 1 & 64 & 56 & RC2-2 & Conv & 7 & 1 & 64 & 56\\
RP2 & MPool & 3 & 2 & 64 & 28 & RP2-1& MPool & 3 & 2 & 64 & 28 &RP2-2& MPool & 3 & 2 & 64 & 28\\
RC3 & Conv & 5 & 1 & 128 & 28 & RC3-1 & Conv& 5 & 1 & 128 & 28 &RC3-2 & Conv& 5 & 1 & 128 & 28\\
RP3 & MPool & 3 & 2 & 128 & 14 & RP3-1 & MPool & 3 & 2 & 128 & 14 & RP3-2 & MPool & 3 & 2 & 128 & 14\\
RC4 & Conv & 3 & 1 & 256 & 14 &RC4-1 & Conv & 3 & 1 & 256 & 14 & RC4-2 & Conv & 3 & 1 & 256 & 14\\
RP4 & MPool & 3 & 2 & 256 & 7 &RP4-1 & MPool & 3& 2& 256 & 7 & RP4-2 & MPool & 3& 2& 256 & 7\\
RC5 & Conv & 3 & 1 & 512 & 7 &RC5-1 & Conv &3 & 1 & 256 & 7 & RC5-2 & Conv &3 & 1 & 256 & 7\\
RP5 & APool & 7 & 1 & 512 & 1 &RP5-1 & APool & 3 & 2 & 256 & 1 & RP5-2 & APool & 3 & 2 & 256 & 1\\
RC6 & Conv & 1 & 1 & 512 & 1 & RP5-1-2 & Concat & $\cdot$ & $\cdot$ & 512 & 1 & RP5-2-1 & Concat & $\cdot$ & $\cdot$ & 512 & 1 \\
RC7 & Conv & 1 & 1 & 512 & 1 & RC6-1 & Conv & 1 & 1 & 512 & 1 & RC6-2 & Conv & 1 & 1 & 512 & 1\\
RC8 & Conv & 1 & 1 & 18 & 1   & RC7-1 & Conv & 1& 1 & 512 & 1 & RC7-2 & Conv & 1& 1 & 512 & 1\\
       &           &    &     &     &      &   RC8-1 & Conv & 1 & 1 & 9 & 1 & RC8-2 & Conv & 1 & 1 & 9 & 1\\
\hline
\end{tabular}
\label{tab:rect}
\end{table*}

Table~\ref{tab:rect} shows the CNN architecture details of the rectification network with both early and late fusions. For the case of the early fusion, $\mathcal{S}_I$ represents the channel-wise concatenation of the two input images $\mathcal{I}_1$ and $\mathcal{I}_2$. The output of ``RC8" is a 18D vector that is split into two 9D vectors to represent $H_1$ and $H_2$ that are fed into the geometric transformation layers. For the case of the late fusion, $\mathcal{I}_1$ and $\mathcal{I}_2$ are processed separately by two convolution towers sharing weights (thus the same convolution tower actually) and their CNN features are fused by the channel-wise concatenation. For the convolution tower for $\mathcal{I}_1$, ``RP5-1-2" is the channel-wise concatenation in the order of the output of ``RP5-1" and output of ``RP5-2". Similarly, for the convolution tower for $\mathcal{I}_2$, ``RP5-2-1" is the channel-wise concatenation in the order of the output of ``RP5-2" and output of ``RP5-1". By further processing these concatenated features, each convolution tower outputs a 9-D homography vector at the end (``RC8-1" and ``RC8-2").

\subsection*{A.2. Encoder-decoder network}

\begin{table*}[]
\centering
\caption{CNN architecture details of the encoder of Deep View Morphing. All convolution layers are followed by ReLU. $k$: kernel size ($k \times k$). $s$: stride in both horizontal and vertical directions. $c$: number of output channels. $d$: output spatial dimension ($d \times d$). Conv: convolution. Deconv: deconvolution. MPool: max pooling. Concat: channel-wise concatenation.}
\label{tab:cnn_details}
\footnotesize
\begin{tabular}{cccccc|ccccccccccccccccc}
\\
\hline
\multicolumn{6}{c|}{{\bf Encoder with early fusion}} &  \multicolumn{12}{c}{{\bf Encoder with late fusion}}\\
\hline
Name & Type & $k$  & $s$ & $c$ & $d$ & Name & Type & $k$ &$s$ & $c$ & $d$ & Name & Type & $k$ &$s$ & $c$ & $d$ &\\
\hline
$\mathcal{S}_R$ & Input & $\cdot$ & $\cdot$ & 6 & 224 & $\mathcal{R}_1$ & Input & $\cdot$ & $\cdot$ &3 & 224& $\mathcal{R}_2$ & Input & $\cdot$ & $\cdot$ &3 & 224\\
EC1 & Conv & 9 & 1 & 32 & 224 & EC1-1 & Conv& 9 & 1 & 32 & 224 & EC1-2 & Conv& 9 & 1 & 32 & 224\\
EP1 & MPool & 3 & 2 & 32 & 112 & EP1-1& MPool& 3 & 2 & 32 & 112 & EP1-2& MPool& 3 & 2 & 32 & 112\\
EC2 & Conv &7 & 1 & 64 & 112 & EC2-1 & Conv & 7 & 1 & 64 & 112 & EC2-2 & Conv & 7 & 1 & 64 & 112\\
EP2 & MPool & 3 & 2 & 64 & 56 &EP2-1& MPool & 3 & 2 & 64 & 56 &EP2-2& MPool & 3 & 2 & 64 & 56\\
EC3 & Conv & 5 & 1 & 128 & 56 &EC3-1 & Conv& 5 & 1 & 128 & 56 &EC3-2 & Conv& 5 & 1 & 128 & 56\\
EP3 & MPool & 3 & 2 & 128 & 28 & EP3-1 & MPool & 3 & 2 & 128 & 28 & EP3-2 & MPool & 3 & 2 & 128 & 28\\
EC4 & Conv & 3 & 1 & 256 & 28 & EC4-1 & Conv & 3 & 1 & 256 & 28 & EC4-2 & Conv & 3 & 1 & 256 & 28\\
EP4 & MPool & 3 & 2 & 256 & 14 & EP4-1 & MPool & 3& 2& 256 & 14 & EP4-2 & MPool & 3& 2& 256 & 14\\
EC5 & Conv & 3 & 1 & 512 & 14 & EC5-1 & Conv &3 & 1 & 512 & 14 & EC5-2 & Conv &3 & 1 & 512 & 14\\
EP5 & MPool & 3 & 2 & 512 & 7 & EP5-1 & MPool & 3 & 2 & 512 & 7 & EP5-2 & MPool & 3 & 2 & 512 & 7\\
EC6 & Conv & 1 & 1 & 1K & 7 & EC6-1 & Conv & 1 & 1 & 512 & 7 & EC6-2 & Conv & 1 & 1 & 512 & 7 \\
 &  & &  & & &  EC6-1-2 & Concat & 1 & 1 & 1K & 7\\
 &  & &  & & &  EC5-1-2 & Concat & 1 & 1 & 1K & 14\\
 &  & &  & & &  EC4-1-2 & Concat & 1 & 1 & 512 & 28\\
 &  & &  & & &  EC3-1-2 & Concat & 1 & 1 & 256 & 56\\
\hline
\end{tabular}
\label{tab:encoder}
\end{table*}

Table~\ref{tab:encoder} shows the CNN architecture details of the encoder with both early and late fusions. For the case of the early fusion, $\mathcal{S}_R$ represents the channel-wise concatenation of the rectified pair $\mathcal{R}_1$ and $\mathcal{R}_2$. The output of ``EC6" is the CNN features that are fed into the decoders. For the case of the late fusion, $\mathcal{R}_1$ and $\mathcal{R}_2$ are processed separately by two encoder towers sharing weights (thus the same encoder tower actually) and their CNN features are fused by the channel-wise concatenation (``EC6-1-2"). ``EC3-1-2", ``EC4-1-2", ``EC5-1-2" are the channel-wise concatenation of CNN features from the lower convolution layers that are used by the visibility decoder. 

\begin{table*}[]
\centering
\caption{CNN architecture details of the correspondence decoder and visibility decoder of Deep View Morphing. All convolution and deconvolution layers are followed by ReLU except for ``CC3" and ``VC3" whose output is the dense correspondences $\mathcal{C}$ and features for the blending masks $\mathcal{M}_1$ and $\mathcal{M}_2$. $k$: kernel size ($k \times k$). $s$: stride in both horizontal and vertical directions. $c$: number of output channels. $d$: output spatial dimension ($d \times d$). Conv: convolution. Deconv: deconvolution. Concat: channel-wise concatenation.}
\label{tab:cnn_details}
\footnotesize
\begin{tabular}{cccccc|ccccccccccccccccc}
\\
\hline
\multicolumn{6}{c|}{{\bf Correspondence decoder}}   & \multicolumn{6}{c}{{\bf Visibility decoder}}\\
\hline
Name & Type & $k$  & $s$ & $c$ & $d$ &Name & Type & $k$ &$s$ & $c$ & $d$\\
\hline
EC6 (or EC6-1-2) & Input & $\cdot$ & $\cdot$ & 1K & 7  & EC6 (or EC6-1-2) & Input & $\cdot$ & $\cdot$ &1K & 7\\
CC1 & Conv & 1 & 1 & 2K & 7 &  VC1 & Conv& 1 & 1 & 1K & 7 \\
CC2 & Conv & 1 & 1 & 2K & 7 &  VC2& Conv& 1 & 1 & 1K & 7 \\
CD1 & Deconv &4 & 2 & 768 & 14  & VD1 & Deconv & 4 & 2 & 512 & 14 \\
CD1-EC5-feature & Concat & $\cdot$ & $\cdot$ & 1K & 14 &  VD2 & Deconv & 4 & 2 & 256 & 28 \\
CD2 & Deconv &4 & 2 & 384 & 28 & VD3 & Deconv & 4 & 2 & 128 & 56 \\
CD2-EC4-feature & Concat & $\cdot$ & $\cdot$ & 512 & 28 & VD4 & Deconv & 4 & 2 & 64 & 112\\
CD3 & Deconv &4 & 2 & 192 & 56 &VD5 & Deconv & 4 & 2 & 32 & 224 \\
CD3-EC3-feature & Concat & $\cdot$ & $\cdot$ & 256 & 56 & VC3 & Conv & 3 & 1 & 1 & 224\\
CD4 & Deconv &4 & 2 & 128 & 112 &  VC3-sig & Sigmoid & $\cdot$ & $\cdot$ & 1 & 224\\
CD5 & Deconv &4 & 2 & 64 & 224 \\
CC3 & Conv & 3 & 1 & 1 & 224\\
\hline
\end{tabular}
\label{tab:decoder}
\end{table*}

Table~\ref{tab:decoder} shows the CNN architecture details of the correspondence decoder and visibility decoder. The correspondence decoder first processes the output of ``EC6" or ``EC6-1-2" by the two convolution layers depending on whether the early or late fusion is used in the encoder. Then the five deconvolution layers perform upsampling of the CNN features. A convolution layer at the end (``CC3") finally outputs the 1D dense correspondence $\mathcal{C}$. In order to increase the accuracy of the predicted dense correspondences, we use the CNN features from the lower convolution layers of the encoder together with those from the last convolution layer of the encoder. We obtain ``EC3-feature", ``EC4-feature", and ``EC5-feature" by applying $1\times1$ kernels to the output of ``EC3" or ``EC3-1-2", ``EC4" or ``EC4-1-2", and ``EC5" or ``EC5-1-2", respectively, depending on whether the early or late fusion is used in the encoder. These are then concatenated to the output of ``CD1", ``CD2", and ``CD3" appropriately (``CD1-EC5-feature'', ``CD2-EC4-feature", and ``CD3-EC3-feature"). 

The visibility decoder is basically the same as the correspondence decoder except it uses the smaller number of channels and it does not use the CNN features from the lower convolution layers of the encoder. The output of the last convolution layer (``VC3") is transformed to the blending mask $\mathcal{M}_1$ by the sigmoid function (``VC3-sig"). Another blending mask $\mathcal{M}_2$ is determined by $1 - \mathcal{M}_1$. 

\subsection*{A.3. How to choose from early and late fusions}
We performed experiments to test all possible combinations of the early and late fusions in the rectification and encoder-decoder networks to find the best one for our view synthesis problem. There are four possible cases: (i) early fusions in both the rectification and encoder-decoder networks, (ii) early fusion in the rectification network and late fusion in the encoder-decoder network, (iii) late fusion in the rectification network and early fusion in the encoder-decoder network, and (iv) late fusions in both the rectification and encoder-decoder networks.

\begin{table}[]
\centering
\caption{Mean of MSE by DVM for ``Car" with different combinations of the early (``E") and late (``L") fusions in the rectification network (``R") and encoder-decoder network (``ED").} 
\label{tab:early_late}
\footnotesize
\begin{tabular}{cccc}
\\
\hline
E(R) + E(ED) & E(R) + L(ED) & L(R) + E(ED) & L(R) + L(ED)\\ 
\hline  48.22 & {\bf 44.70} &  49.59 & 46.18\\
\hline
\end{tabular}
\end{table}

%\begin{table}[]
%\centering
%\caption{Mean of MSE by DVM for ``Car" with different combinations of the early  and late fusions in the rectification network (``R") and encoder-decoder network (``ED").} 
%\label{tab:early_late}
%\small
%\begin{tabular}{cc}
%\\
%\hline
%Early (R) + Early (ED) & Early (R) + Late (ED)\\ 
%\hline  48.22 & {\bf 44.70}\\
%\hline
%\hline
%Late (R) + Early (ED) & Late (R) + Late (ED)\\
%\hline 49.59 & 46.18 \\
%\hline
%\end{tabular}
%\end{table}

%\begin{table}[]
%\centering
%\caption{Mean and standard deviation of MSE by DVM for ``Car" with different combinations of the early and late fusions in the rectification network (``R") and encoder-decoder network (``ED").} 
%\label{tab:early_late}
%\small
%\begin{tabular}{cc}
%\\
%\hline
%Early (R) + Early (ED) & Early (R) + Late (ED)\\ 
%\hline  48.22 ($\pm$45.26) & {\bf 44.70 ($\pm$42.69)}\\
%\hline
%\hline
%Late (R) + Early (ED) & Late (R) + Late (ED)\\
%\hline 49.59 ($\pm$46.43) & 46.18 ($\pm$44.01)\\
%\hline
%\end{tabular}
%\end{table}

Table.~\ref{tab:early_late} shows the mean of MSE for ``Car" by DVM  with the four cases considered. Although it is difficult to draw any generalizable conclusions from Tab.~\ref{tab:early_late}, at least it explains our CNN architecture design choice: the early fusion in the rectification network and late fusion in the encoder-network yields the best results for our view synthesis problem.\\ 

\subsection*{A.4. Other implementation details}
We set the spatial dimensions of $\mathcal{I}_1$, $\mathcal{I}_2$, $\mathcal{R}_1$, $\mathcal{R}_2$, and $\mathcal{R}_{\alpha=0.5}$ all to be $224 \times 224$. We normalize each channel of $\mathcal{I}_1$ and $\mathcal{I}_2$ to the range of $-0.5$ to $0.5$ by global shift by 128 and division by 255. We use 2D bilinear interpolation for the geometric transformation layers in the rectification network and 1D linear interpolation for the sampling layers in the view morphing network.

\section*{Appendix B. Differentiation of geometric transformations}
The geometric transformation of images with homographies in the rectification network can be split into two steps: (i) pixel coordinate transformation by homographies and (ii) pixel value sampling with the transformed pixel coordinates by 2D bilinear interpolation. We show how to obtain gradients of each step below.

With a homogrpahy $H= \left[ \begin{smallmatrix} h_1 & h_2 & h_3 \\ h_4 & h_5 & h_6 \\ h_7 & h_8 & h_9 \end{smallmatrix} \right]$, the pixel coordinates $p =(p_x, p_y)^\top$ are transformed to $r = (r_x, r_y)^\top$ as 
\begin{equation}
\left[ \begin{matrix} r_x \\ r_y \end{matrix} \right] = \left[ \begin{matrix} \frac{h_1 p_x + h_2 p_y + h_3}{h_7 p_x + h_8 p_y + h_9}  \\ \frac{h_4 p_x + h_5 p_y + h_6}{h_7 p_x + h_8 p_y + h_9} \end{matrix} \right].
\label{eqn:homography} 
\end{equation}
We can obtain $2\times9$ Jacobian matrix $J  = \left[ \begin{smallmatrix} \frac{\partial r_x}{\partial h_i}  \frac{\partial r_y}{\partial h_i} \end{smallmatrix}  \right]^\top$ of this coordinate transformation with respect to $h_i$, each element of the homogrpahy matrix, as
\begin{equation}
\small
\left[
\begin{matrix}
\frac{p_x}{p_3}  & \frac{p_y}{p_3} & \frac{1}{p_3} & 0 & 0 & 0 & -\frac{p_1}{p_3^2}p_x & -\frac{p_1}{p_3^2}p_y & -\frac{p_1}{p_3^2} \\
        0 & 0 & 0  & \frac{p_x}{p_3} & \frac{p_y}{p_3} & \frac{1}{p_3} & -\frac{p_2}{p_3^2}p_x & -\frac{p_2}{p_3^2}p_y & -\frac{p_2}{p_3^2}
\end{matrix}
\right],
\end{equation}
where $p_1 = h_1p_x + h_2p_y + h_3$, $p_2 = h_4p_x + h_5p_y + h_6$, and $p_3 = h_7p_x + h_8p_y + h_9$. 

As shown in \cite{STN}, 2D bilinear interpolation to sample pixel values of an image $\mathcal{I}$ at the transformed pixel coordinates $r$ can be expressed as
\begin{equation}
\begin{split}
\mathcal{I}^c(r_x, r_y) = & \sum_{n=\lfloor r_y \rfloor}^{\lceil r_y \rceil}  \sum_{m=\lfloor r_x \rfloor}^{\lceil r_x \rceil} \mathcal{I}^c(m,n) 
\\  
&  \cdot (1 - | r_x - m |) \cdot (1 - | r_y - n |), 
\end{split}
\label{eqn:bilinear}
\end{equation}
where $\mathcal{I}^c$ represents each color channel of $\mathcal{I}$. This 2D bilinear interpolation can be reduced to 1D linear interpolation that is used in the sampling layer of the view morphing network if we only consider components related to $r_x$.

The gradients of the 2D bilinear interpolation in (\ref{eqn:bilinear}) with respect to $r=(r_x, r_y)^\top$ are
\begin{equation}
\begin{split}
\frac{\partial \mathcal{I}^c(r_x, r_y)}{\partial r_x}  = &  \sum_{n=\lfloor r_y \rfloor}^{\lceil r_y \rceil}  \sum_{m=\lfloor r_x \rfloor}^{\lceil r_x \rceil} \mathcal{I}^c(m,n) 
\\
& \cdot (1 - |r_y - n|) \cdot 
\begin{cases}
\text{ }\text{ }\text{ } 1 \textrm{ }\text{ if }\text{ } m \geq r_x \\
-1 \textrm{ }\text{ if }\text{ } m <  r_x \\
\end{cases},
\\
\frac{\partial \mathcal{I}^c(r_x, r_y)}{\partial r_y}  = &  \sum_{n=\lfloor r_y \rfloor}^{\lceil r_y \rceil}  \sum_{m=\lfloor r_x \rfloor}^{\lceil r_x \rceil} \mathcal{I}^c(m,n) 
\\
& \cdot (1 - |r_x- m|) \cdot 
\begin{cases}
\text{ }\text{ }\text{ } 1 \textrm{ }\text{ if }\text{ } n \geq r_y \\
-1 \textrm{ }\text{ if }\text{ } n <  r_y \\
\end{cases},
\end{split}
\label{eqn:bilinear_gradient}
\end{equation}
while those with respect to $\mathcal{I}^c(m,n)$ are given by
\begin{equation}
\frac{\partial \mathcal{I}^c(r_x,r_y)}{\partial \mathcal{I}^c(m,n)} = \max(0, 1 - |r_x - m|) \cdot \max(0, 1 - |r_y - n|).
\label{eqn:bilinear_gradient_image}
\end{equation}
Note that the gradients in (\ref{eqn:bilinear_gradient_image}) that are reduced to the 1D case are used in the view morphing network for error backpropagation while they are not used in the rectification network because the input images are fixed.

\section*{Appendix C. More experimental results}

\subsection*{C.1. ShapeNet}

\begin{table}[]
\centering
\caption{Numbers of 3D models and triplets for ``Car", ``Chair", ``Airplane", and ``Vessel" of ShapeNet.}
\label{tab:shapenet}
\footnotesize
\begin{tabular}{cccccc}
\\
\hline
  & & Car & Chair & Airplane & Vessel\\
\hline
\multirow{2}{*}{Train} & Models  & 5,997 & 5,422 & 3,236 & 1,551\\
& Triplets & 3,454,272 & 3,123,072 & 1,863,936 & 893,376\\
\hline
\multirow{2}{*}{Test} & Models & 1,499 & 1,356 & 809 & 388\\
& Triplets & 200,000 & 200,000 & 200,000 & 200,000\\
\hline
\end{tabular}
\end{table}

\begin{table}[]
\centering
\caption{Numbers of 3D models and test triplets for the unseen ``Motorcycle", ``Laptop", ``Clock", and ``Bookshelf" of ShapeNet.}
\label{tab:shapenet_unseen}
\footnotesize
\begin{tabular}{ccccc}
\\
\hline
  &  Motorcycle & Laptop & Clock & Bookshelf\\
\hline
Models  & 337 & 460 & 655 & 466\\
Triplets & 194,112 & 200,000 &200,000 & 200,000\\
\hline
\end{tabular}
\end{table}

{\setlength{\parindent}{0cm}{\bf More details of training and test data.}}
Table~\ref{tab:shapenet} shows the details of the ShapeNet training and test data of ``Car", ``Chair", ``Airplane", and ``Vessel" used for evaluating the category-specific training results while Tab.~\ref{tab:shapenet_unseen} shows the details of the unseen ShapeNet test data of ``Motorcycle", ``Laptop", ``Clock", and ``Bookshelf" used for evaluating the category-agnostic training results.\\

\begin{figure*}
\centering
\includegraphics[width=6.2in]{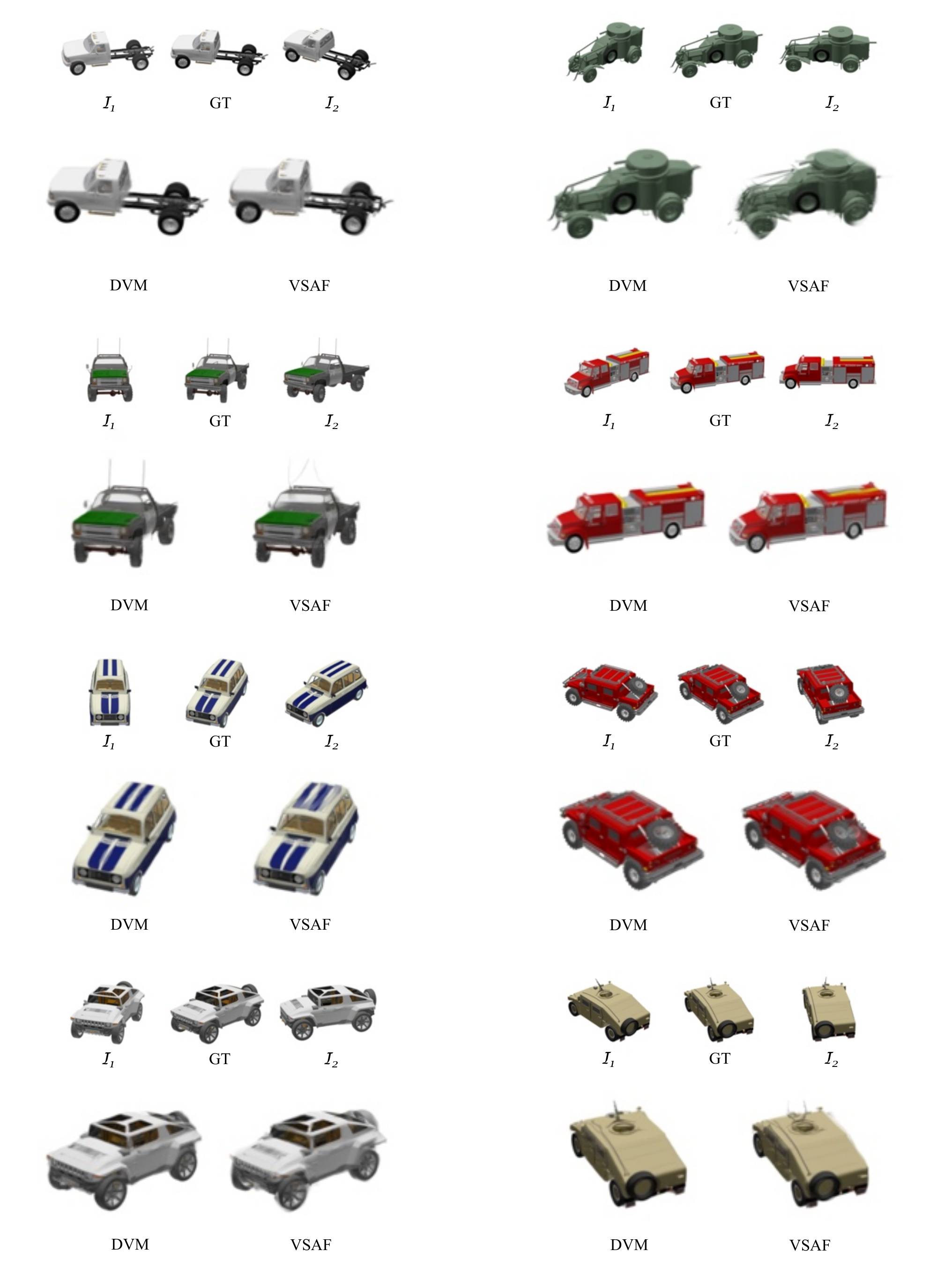}
\caption{Comparisons of view synthesis results by DVM and VSAF on test samples of ``Car" of ShapeNet. Two input images are shown on the left and right sides of the ground truth image (``GT").}
\label{fig:results_shapenet_car}
\end{figure*}

\begin{figure*}
\centering
\includegraphics[width=6.2in]{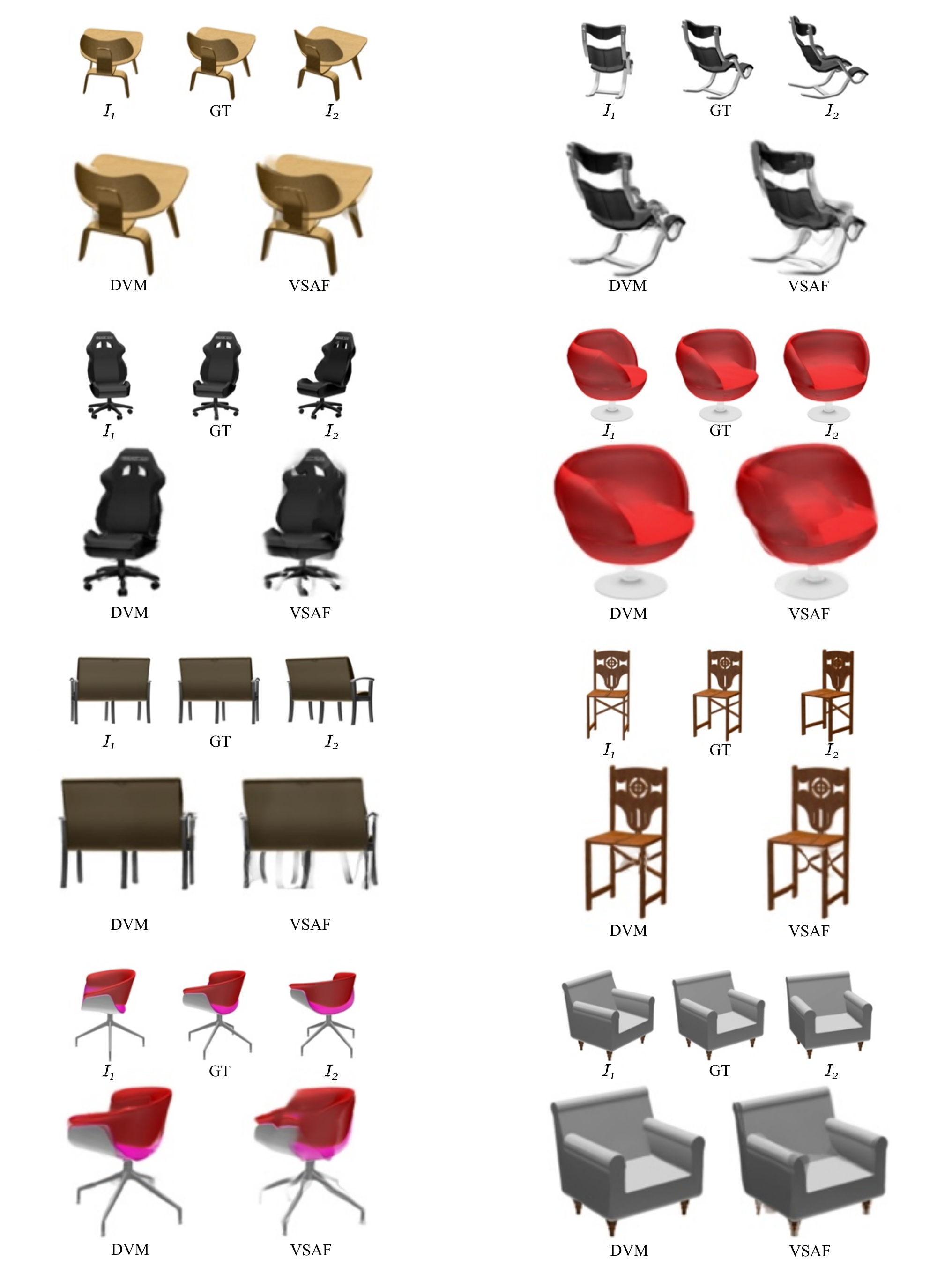}
\caption{Comparisons of view synthesis results by DVM and VSAF on test samples of ``Chair" of ShapeNet. Two input images are shown on the left and right sides of the ground truth image (``GT").}
\label{fig:results_shapenet_chair}
\end{figure*}

\begin{figure*}
\centering
\includegraphics[width=6.2in]{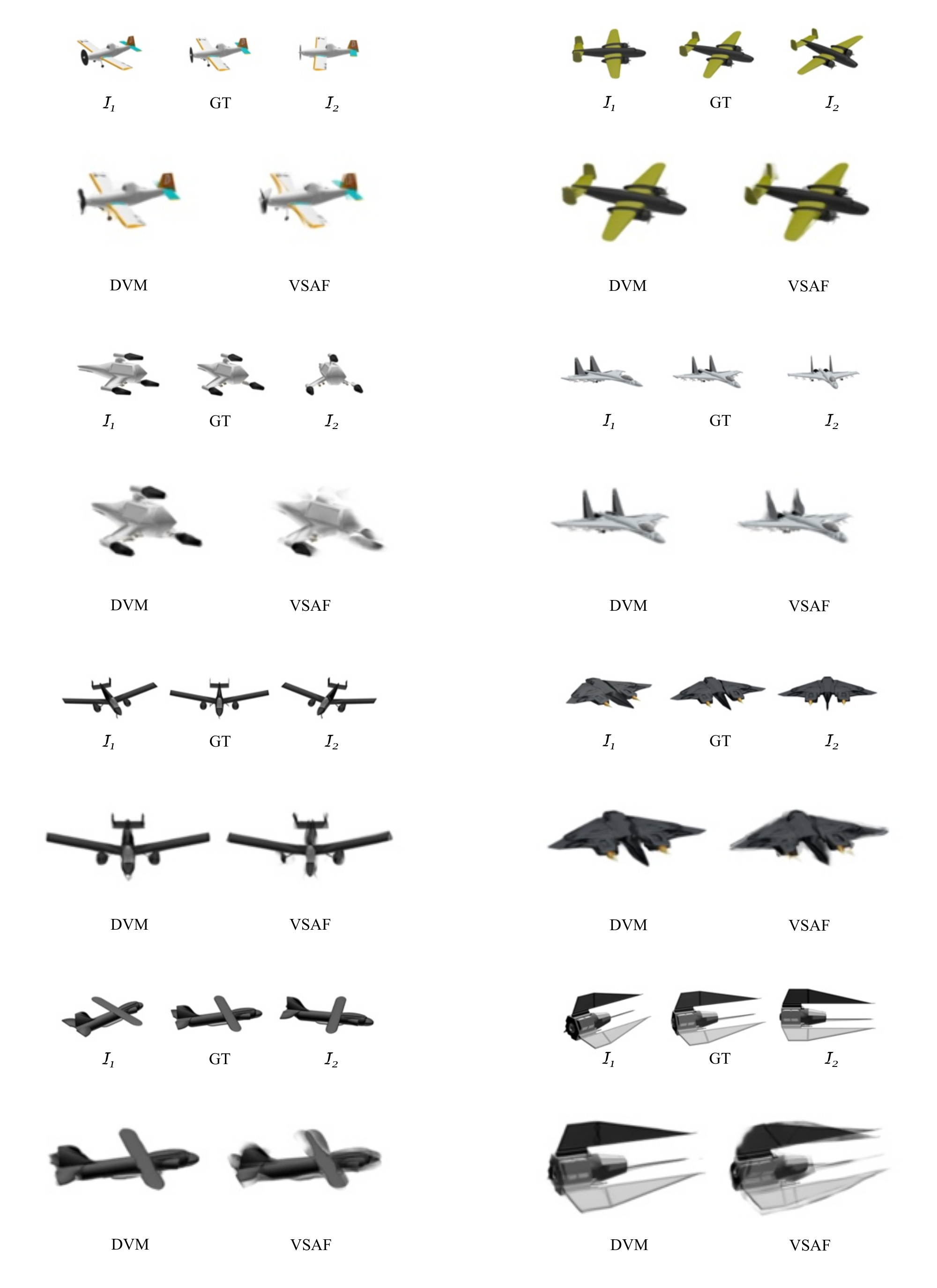}
\caption{Comparisons of view synthesis results by DVM and VSAF on test samples of ``Airplane" of ShapeNet. Two input images are shown on the left and right sides of the ground truth image (``GT").}
\label{fig:results_shapenet_chair}
\end{figure*}

\begin{figure*}
\centering
\includegraphics[width=6.2in]{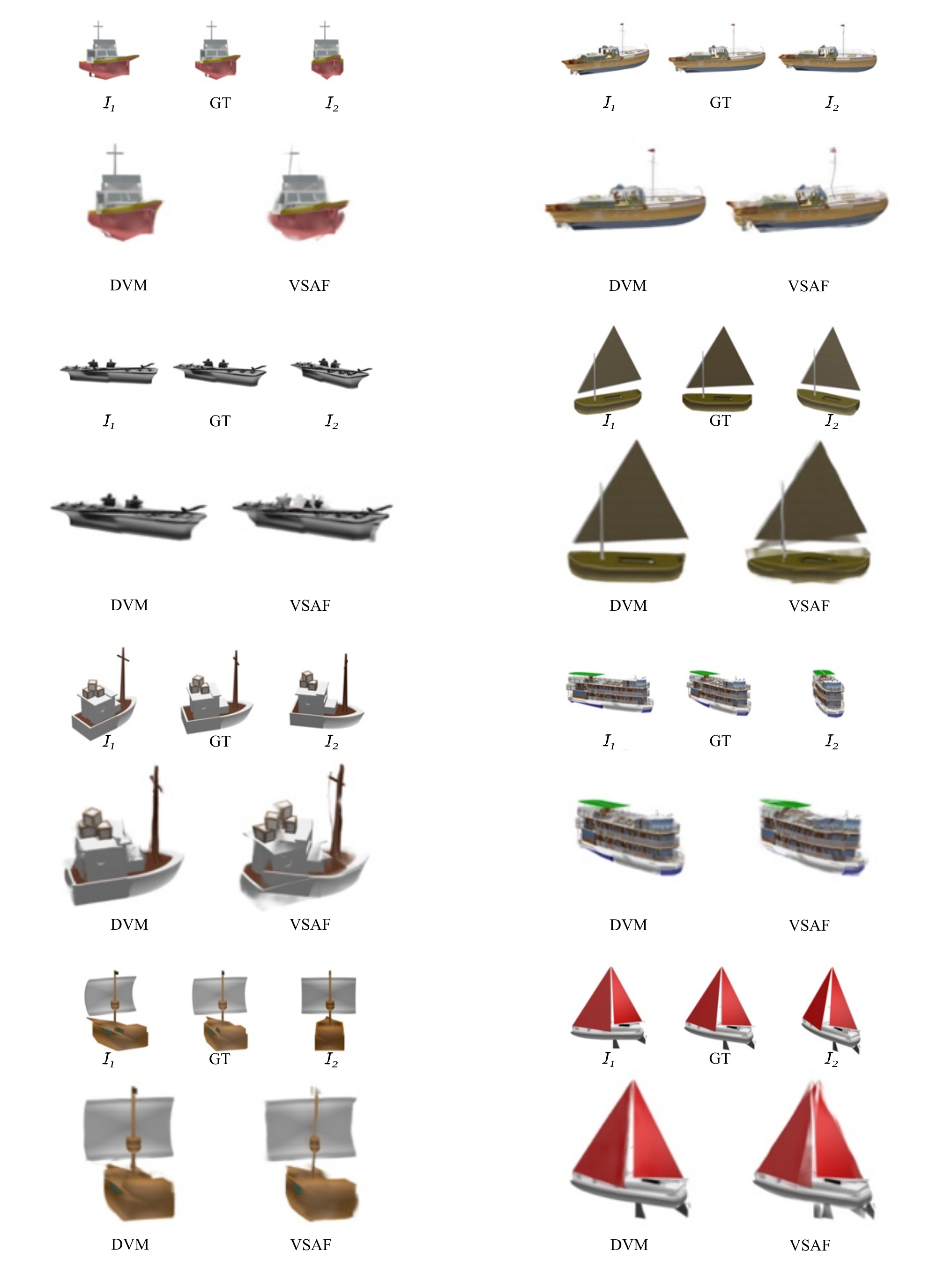}
\caption{Comparisons of view synthesis results by DVM and VSAF on test samples of ``Vessel" of ShapeNet. Two input images are shown on the left and right sides of the ground truth image (``GT").}
\label{fig:results_shapenet_vessel}
\end{figure*}

\begin{figure*}
\centering
\includegraphics[width=4.3in]{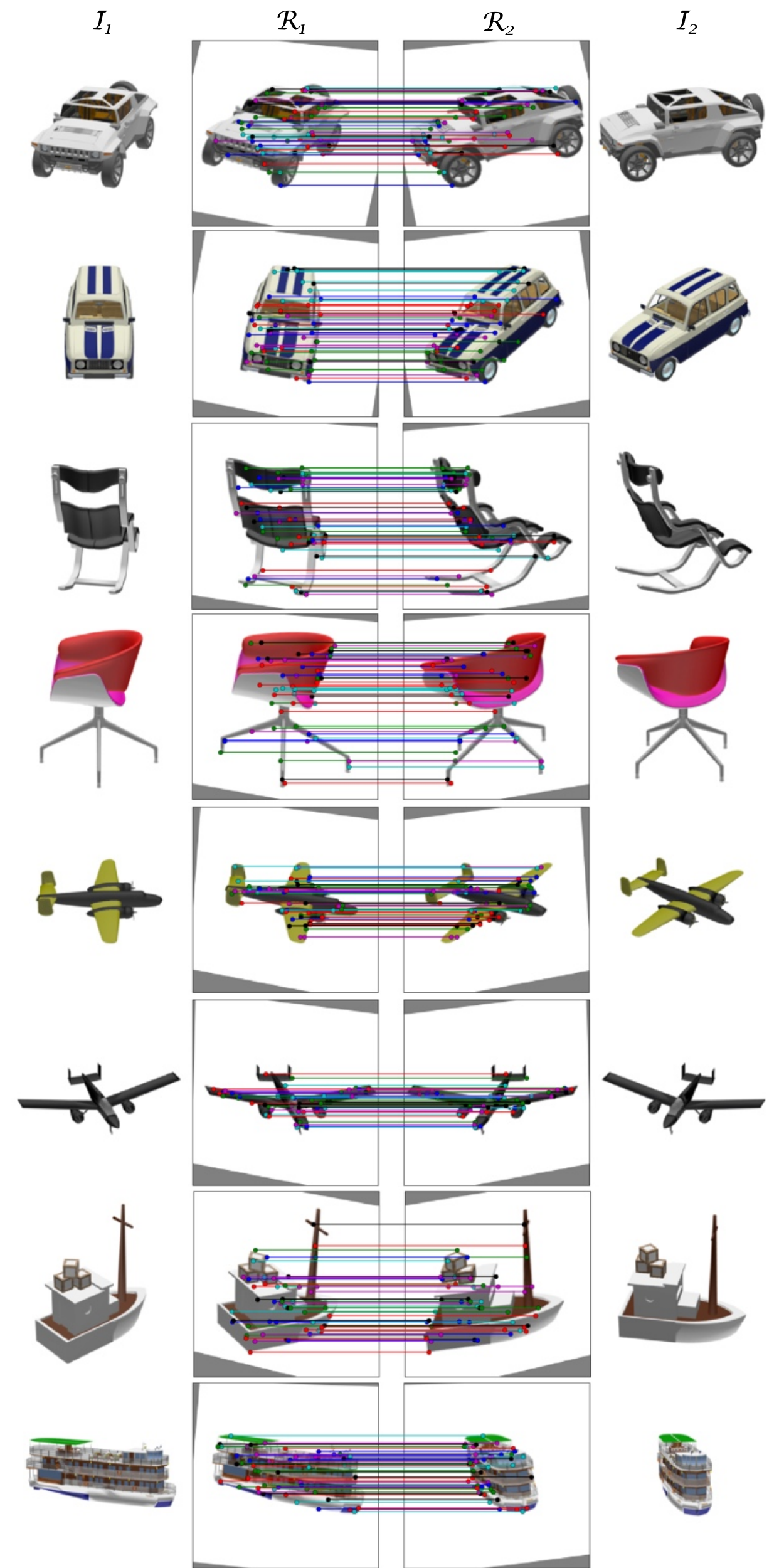}
\caption{Examples of rectification results and dense correspondences obtained by DVM trained in a category-specific way on the test input images of ``Car", ``Chair", ``Airplane", and ``Vessel" of ShapeNet.}
\label{fig:corr_shapenet_seen}
\end{figure*}

\begin{figure*}
\centering
\includegraphics[width=6.2in]{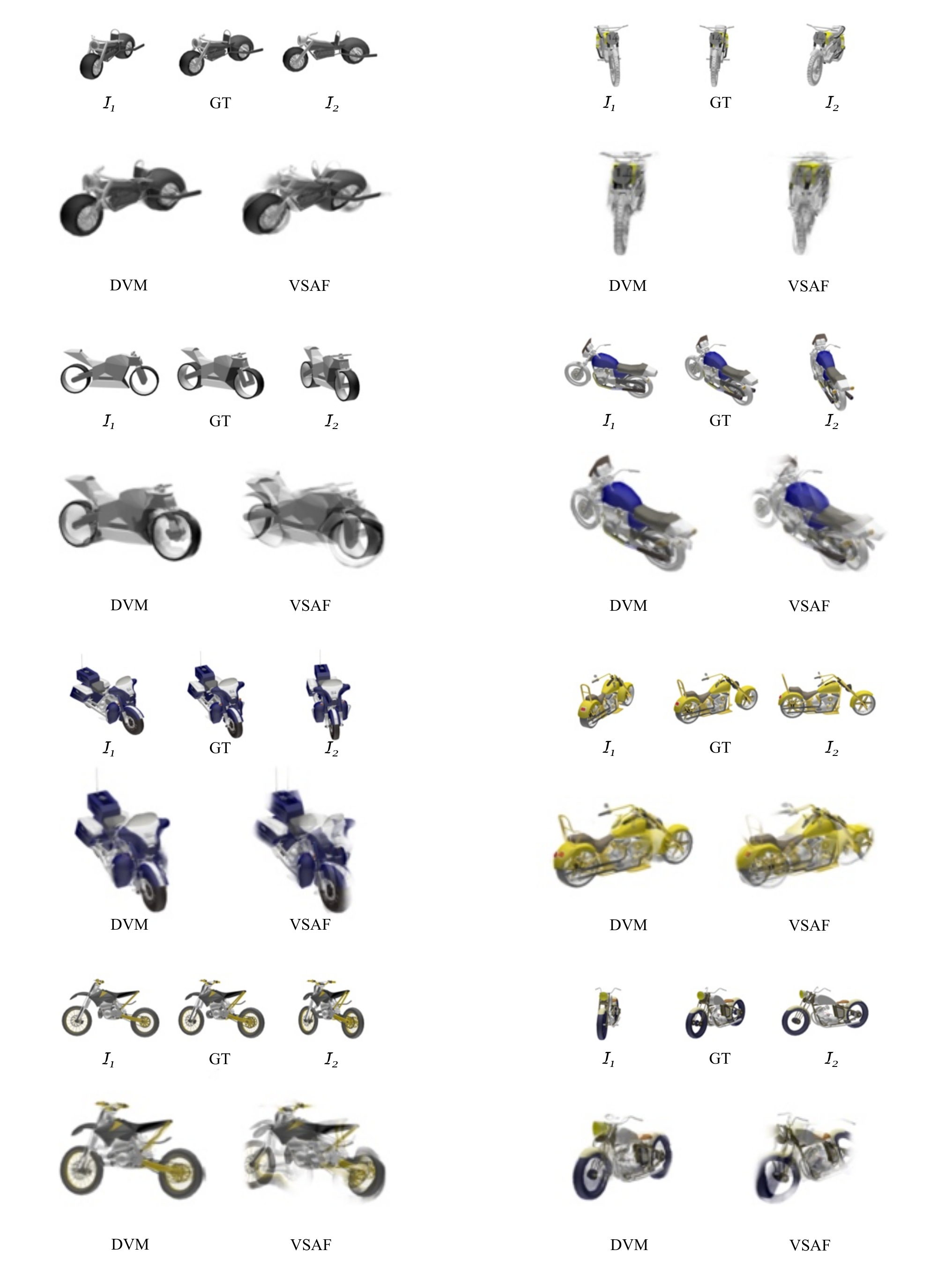}
\caption{Comparisons of view synthesis results by DVM and VSAF on test samples of the unseen ``Motorcycle" of ShapeNet. Two input images are shown on the left and right sides of the ground truth image (``GT").}
\label{fig:results_shapenet_motorcycle}
\end{figure*}

\begin{figure*}
\centering
\includegraphics[width=6.2in]{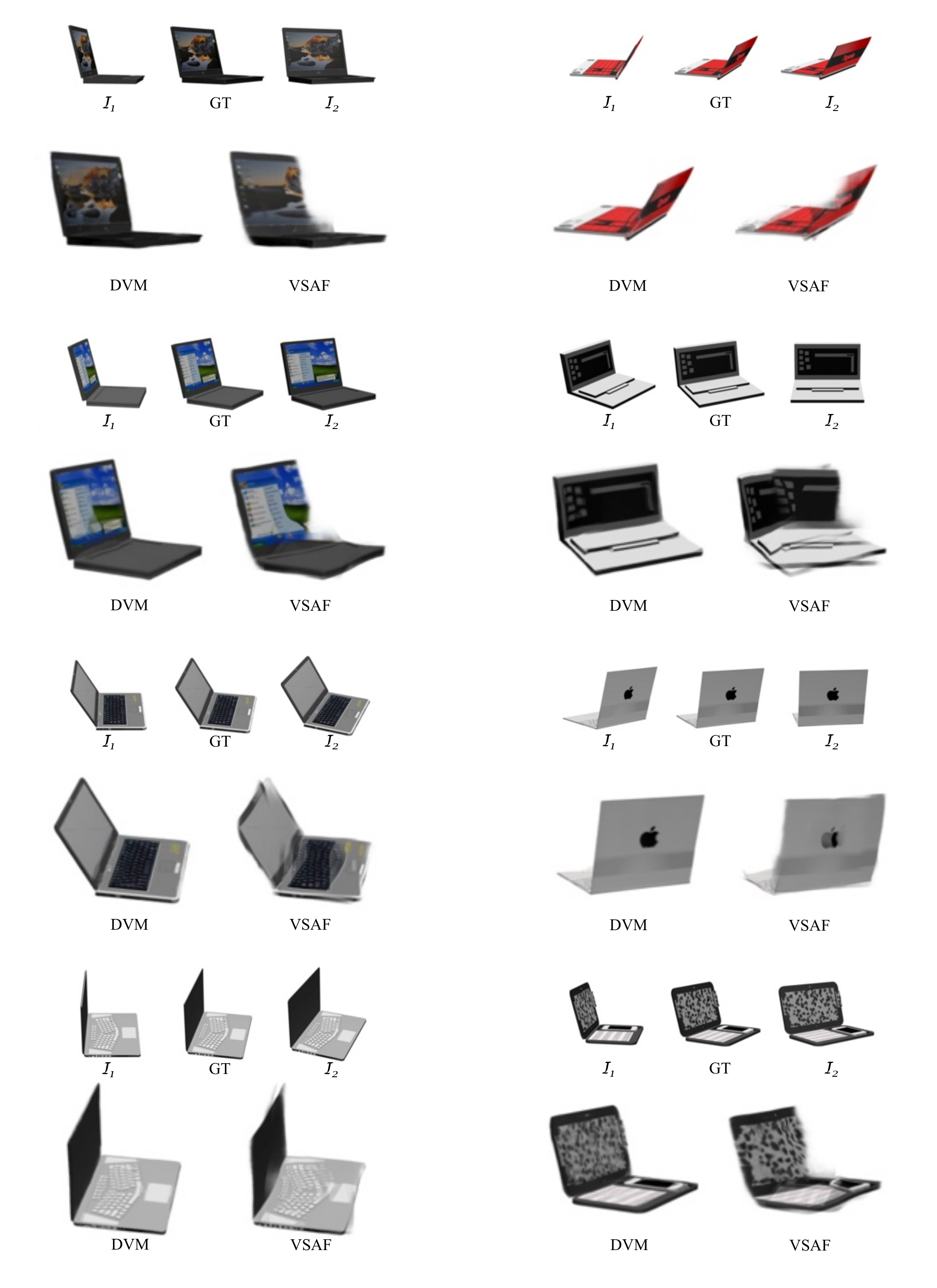}
\caption{Comparisons of view synthesis results by DVM and VSAF on test samples of the unseen ``Laptop" of ShapeNet. Two input images are shown on the left and right sides of the ground truth image (``GT").}
\label{fig:results_shapenet_laptop}
\end{figure*}

\begin{figure*}
\centering
\includegraphics[width=6.2in]{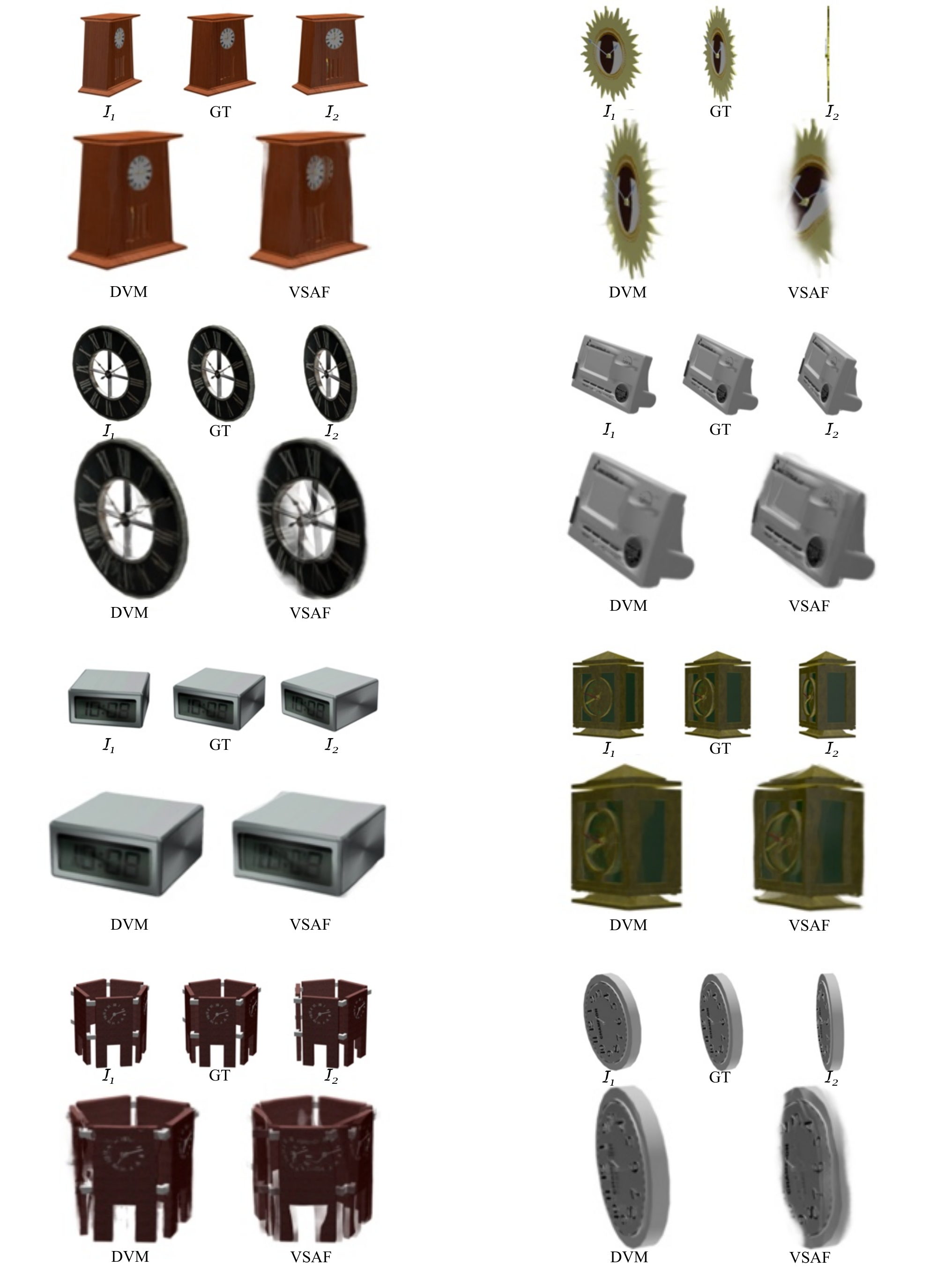}
\caption{Comparisons of view synthesis results by DVM and VSAF on test samples of the unseen ``Clock" of ShapeNet. Two input images are shown on the left and right sides of the ground truth image (``GT").}
\label{fig:results_shapenet_clock}
\end{figure*}

\begin{figure*}
\centering
\includegraphics[width=6.2in]{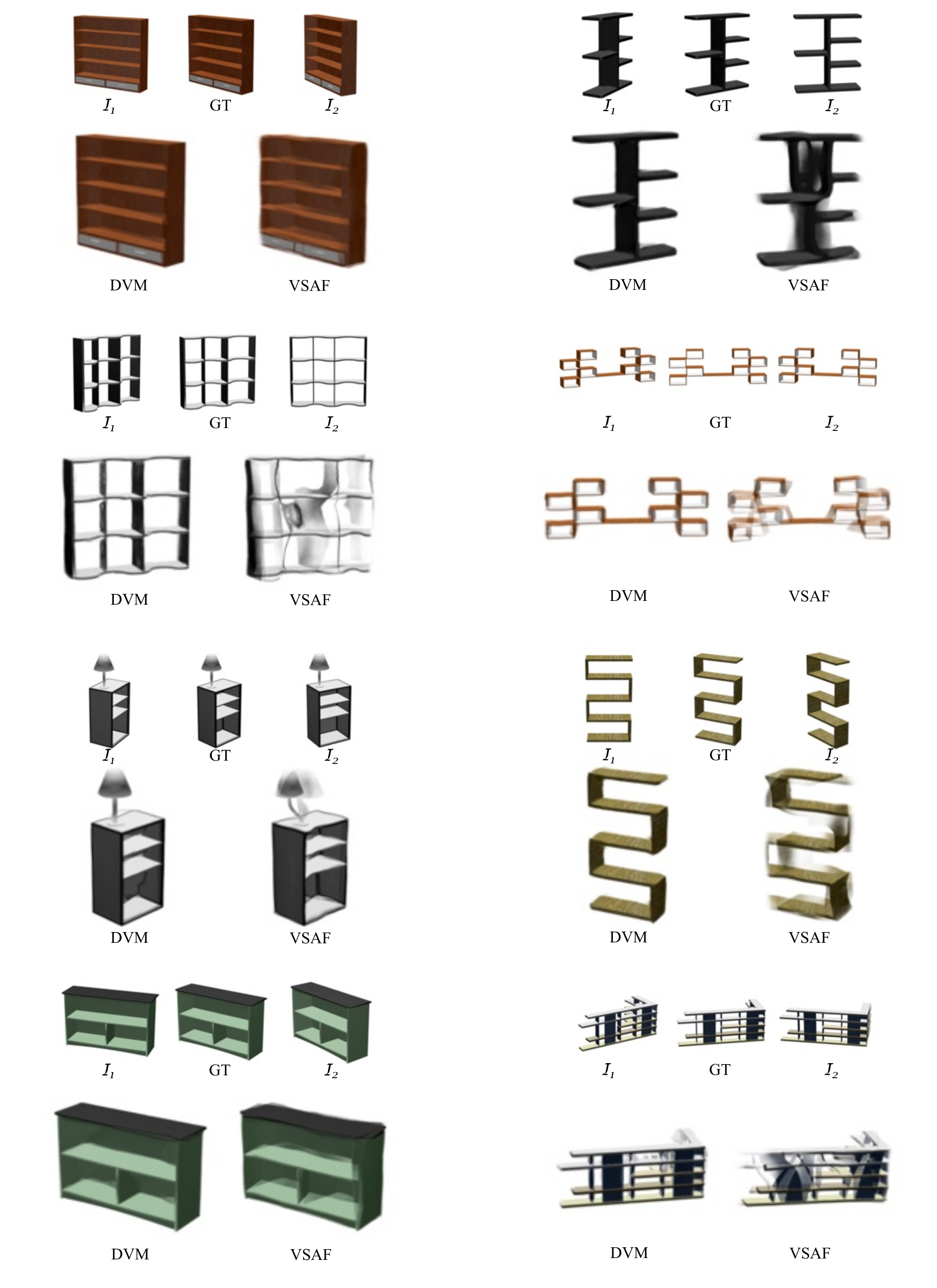}
\caption{Comparisons of view synthesis results by DVM and VSAF on test samples of the unseen ``Bookshelf" of ShapeNet. Two input images are shown on the left and right sides of the ground truth image (``GT").}
\label{fig:results_shapenet_bookshelf}
\end{figure*}

\begin{figure*}
\centering
\includegraphics[width=4.3in]{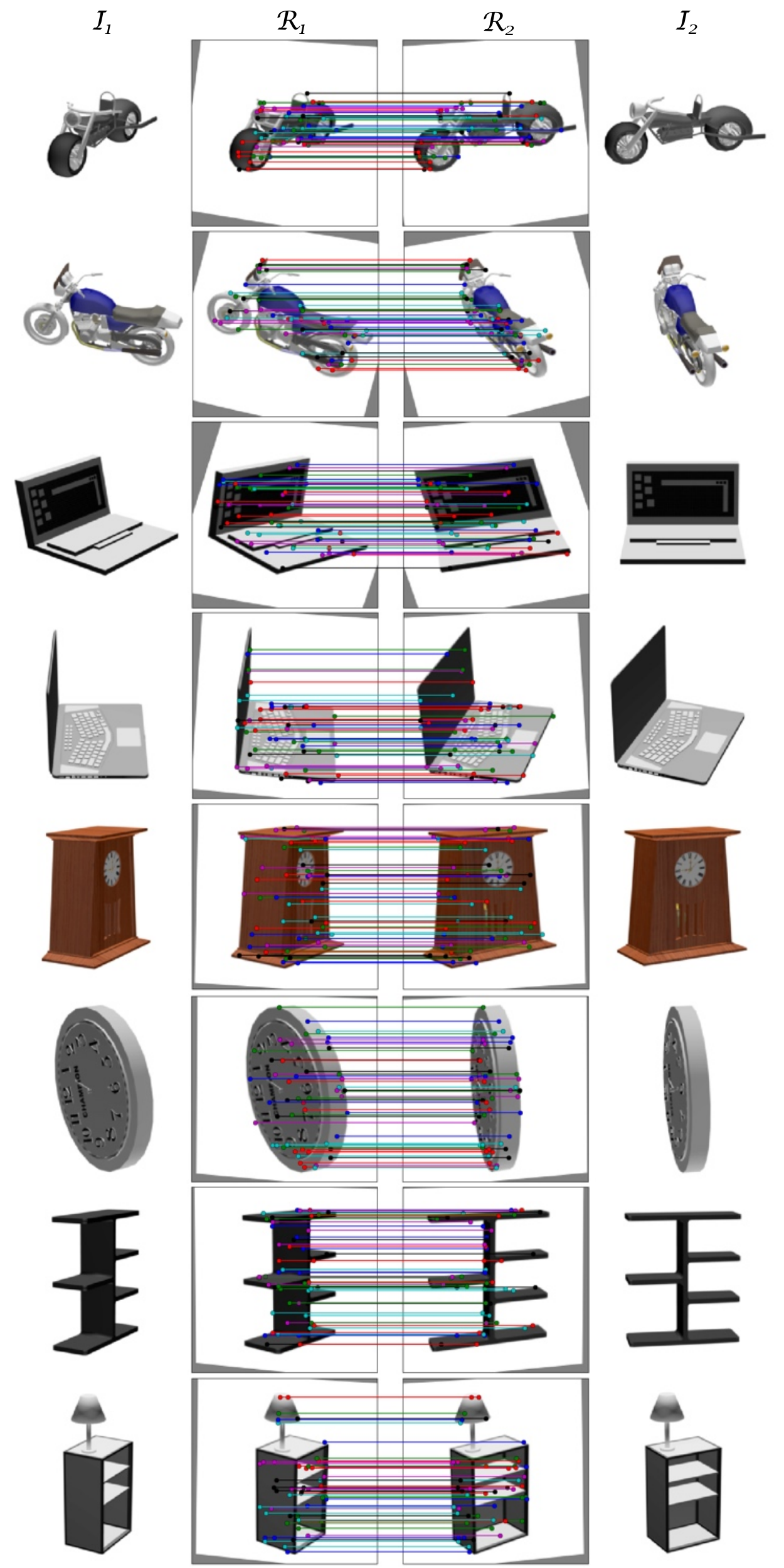}
\caption{Examples of rectification results and dense correspondences obtained by DVM trained in a category-agnostic way on the test input images of the unseen ``Motorcycle", ``Laptop", ``Clock", and ``Bookshelf" of ShapeNet.}
\label{fig:corr_shapenet_unseen}
\end{figure*}

{\setlength{\parindent}{0cm}{\bf More qualitative results.}}
Figure~\ref{fig:results_shapenet_car} to Fig.~\ref{fig:results_shapenet_vessel} show more qualitative comparisons of the view synthesis results on ``Car", ``Chair", ``Airplane", and ``Vessel" of ShapeNet by DVM and VSAF \cite{Zhou_eccv} trained in a category-specific way. Figure~\ref{fig:corr_shapenet_seen} shows more examples of the rectification results and dense correspondence results by DVM.

Figure~\ref{fig:results_shapenet_motorcycle} to Fig.~\ref{fig:results_shapenet_bookshelf} show more qualitative comparisons of the view synthesis results on the unseen ``Motorcycle", ``Laptop", ``Clock", and ``Bookshelf" of ShapeNet by DVM and VSAF trained in a category-agnostic way. Figure~\ref{fig:corr_shapenet_unseen} shows examples of the rectification results and dense correspondence results by DVM on the unseen categories.\\

{\setlength{\parindent}{0cm}{\bf Rectification accuracy.}} The rectification network is trained to rectify $\mathcal{I}_1$ and $\mathcal{I}_2$ so that the middle view of $\mathcal{R}_1$ and $\mathcal{R}_2$ can be directly matched against the desired ground truth middle view $\mathcal{R}_{\textrm{GT}}$. We can measure how successful the rectification is by checking how well aligned the known points in the ground truth middle view are to the corresponding points in the rectified pair $\mathcal{R}_1$ and $\mathcal{R}_2$. 

Let $T_1$, $T_{\textrm{GT}}$, and $T_2$ denote the known camera poses used for rendering a test triplet $\{ \mathcal{I}_1, \mathcal{R}_{\textrm{GT}}, \mathcal{I}_2 \}$. In the camera coordinate frames of $T_{\textrm{GT}}$, we put four lines with end points $l_1^i$ and $l_2^i, i=1,\ldots,4$, as 
\begin{equation}
\begin{split}
& l_1^1 = (-0.1, -0,1, 3.5)^\top, \textrm{ }\textrm{ } l_2^1 = (0.1, -0.1, 3.5)^\top,\\
& l_1^2 = (-0.1, 0.1, 3.5)^\top, \textrm{ }\textrm{ }\textrm{ }\textrm{ }\textrm{ }\textrm{ } l_2^2 = (0.1, 0.1, 3.5)^\top,\\
& l_1^3 = (-0.1, -0.1, 4.5)^\top, \textrm{ }\textrm{ }\textrm{ } l_2^3 = (0.1, -0.1, 4.5)^\top,\\
& l_1^4 = (-0.1, 0.1, 4.5)^\top, \textrm{ }\textrm{ }\textrm{ }\textrm{ }\textrm{ }\textrm{ } l_2^4 = (0.1, 0.1, 4.5)^\top.
\end{split}
\label{eqn:points}
\end{equation}
These lines will be projected onto the image plane of $T_{\textrm{GT}}$ as horizontal lines. Note that the distance from the camera to 3D models in rendering was 4. 

As we know the exact intrinsic camera parameters used for rendering and the relative camera poses of $T_1$ and $T_2$ with respect to $T_{\textrm{GT}}$,  we can obtain the projections of the four lines in (\ref{eqn:points}) onto the image planes of $T_1$ and $T_2$. Before the rectification, these lines will be projected to the image planes of $T_1$ and $T_2$ as slanted. After applying the homographies $H_1$ and $H_2$ predicted by the rectification network to those projected lines of $T_1$ and $T_2$, they will be aligned well to the corresponding projected lines of $T_{\textrm{GT}}$. 

As a measure for the rectification error, we compute the average vertical difference $D$ between the end points of the corresponding projected lines of  $T_{\textrm{GT}}$ and $T_1$ and $T_2$ after the rectification. With $a_0^i$ and $a_1^i$ to represent $y$-components of projections of the end points in (\ref{eqn:points}) onto the image plane of $T_{\textrm{GT}}$, we compute $D$ as 
\begin{equation}
\begin{split}
D = & \frac{1}{16} \left( \sum_{i=1}^4 | a_{0}^i - b_{0}^i | +  | a_{1}^i - b_{1}^i |  \right.\\
& \textrm{ }\textrm{ }\textrm{ }\textrm{ }+ \left. \sum_{i=1}^4 | a_{0}^i - c_{0}^i | +  | a_{1}^i - c_{1}^i | \right),
\end{split}
\label{eqn:D}
\end{equation}
where $b_0^i$ and $b_1^i$ are $y$-components of the corresponding projections onto the image plane of $T_1$ after the rectification and $c_0^i$ and $c_1^i$ are those of $T_2$ also after the rectification.

\begin{table}[]
\centering
\caption{Mean of the rectification error $D$ in (\ref{eqn:D}) by DVM for the ShapeNet test triplets. The numbers in parenthesis are the standard deviations of $D$.}
\label{tab:rectification_accuracy}
\footnotesize
\begin{tabular}{cccc}
\\
\hline
\multicolumn{4}{c}{{\bf Category-specific training}}\\
\hline
Car & Chair & Airplane & Vessel\\
\hline
1.314 ($\pm1.229$) & 1.122 ($\pm1.081$) & 1.294 ($\pm1.213$) & 1.336 ($\pm1.225$)\\
\hline
\hline
\multicolumn{4}{c}{{\bf Category-agnostic training}}\\
\hline
Car & Chair & Airplane & Vessel \\
\hline
1.307 ($\pm1.228$) & 1.138 ($\pm1.091$) & 1.319 ($\pm1.244$) & 1.299 ($\pm1.206$) \\
\hline
Motorcycle & Laptop & Clock & Bookshelf\\
\hline
1.349 ($\pm 1.252$) & 1.169 ($\pm1.094$) & 1.529 ($\pm 1.154$) & 1.224 ($\pm1.107$) \\
\hline
\end{tabular}
\end{table}

Table~\ref{tab:rectification_accuracy} shows the mean of $D$ by DVM on the ShapeNet test triplets. Note that the rectification error by DVM for ``Car", ``Chair", ``Airplane", and ``Vessel" is quite consistent for both category-specific training and category-agnostic training. Similarly to MSE of the view synthesis results, the rectification error for ``Vessel" by the category-agnostic training is decreased by the training data of the other categories. It is significant that the rectification error for the unseen ``Motorcycle", ``Laptop", and ``Bookshelf" are quite similar to that for the seen categories. The rectification error for the unseen ``Clock" is relatively large, which leads to the relatively large MSE of the view synthesis results.\\

\begin{figure}
\centering
\includegraphics[width=2.8in]{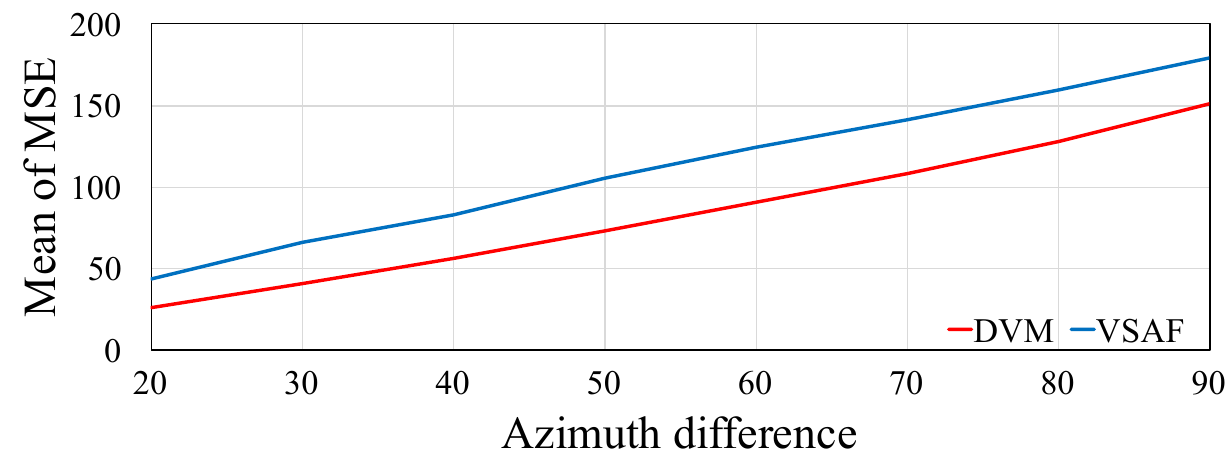}
\caption{Plots of mean of MSE by DVM (red) and VSAF (blue) as a function of azimuth difference $\Delta \phi$  between $\mathcal{I}_1$ and $\mathcal{I}_2$ for ``Car". Here, the azimuth differences are $20^\circ \leq \Delta \phi < 90^\circ$ with $10^\circ$ steps.}
\label{fig:errors_shapenet_larger_vp}
\end{figure}

{\setlength{\parindent}{0cm}{\bf Larger azimuth differences.}} One can argue that VSAF is originally designed to be able to deal with larger azimuth differences. Therefore, we test the performance of DVM and VSAF for azimuth differences up to $90^\circ$. We trained DVM and VSAF using training triplets of ``Car" newly created with $20^\circ \leq \Delta \phi \leq 90^\circ$ with $10^\circ$ steps. We provided VSAF with 16-D one-hot vectors as viewpoint transformation input. 

Figure~\ref{fig:errors_shapenet_larger_vp} shows the mean of MSE by DVM and VSAF on the test triplets of ``Car" with $20^\circ \leq \Delta \phi \leq 90^\circ$ with $10^\circ$ steps. As long as DVM and VSAF are trained using the same data, DVM consistently outperforms VSAF even for larger azimuth differences up to $90^\circ$.

\subsection*{C.2. Multi-PIE}

\begin{figure*}
\centering
\includegraphics[width=6in]{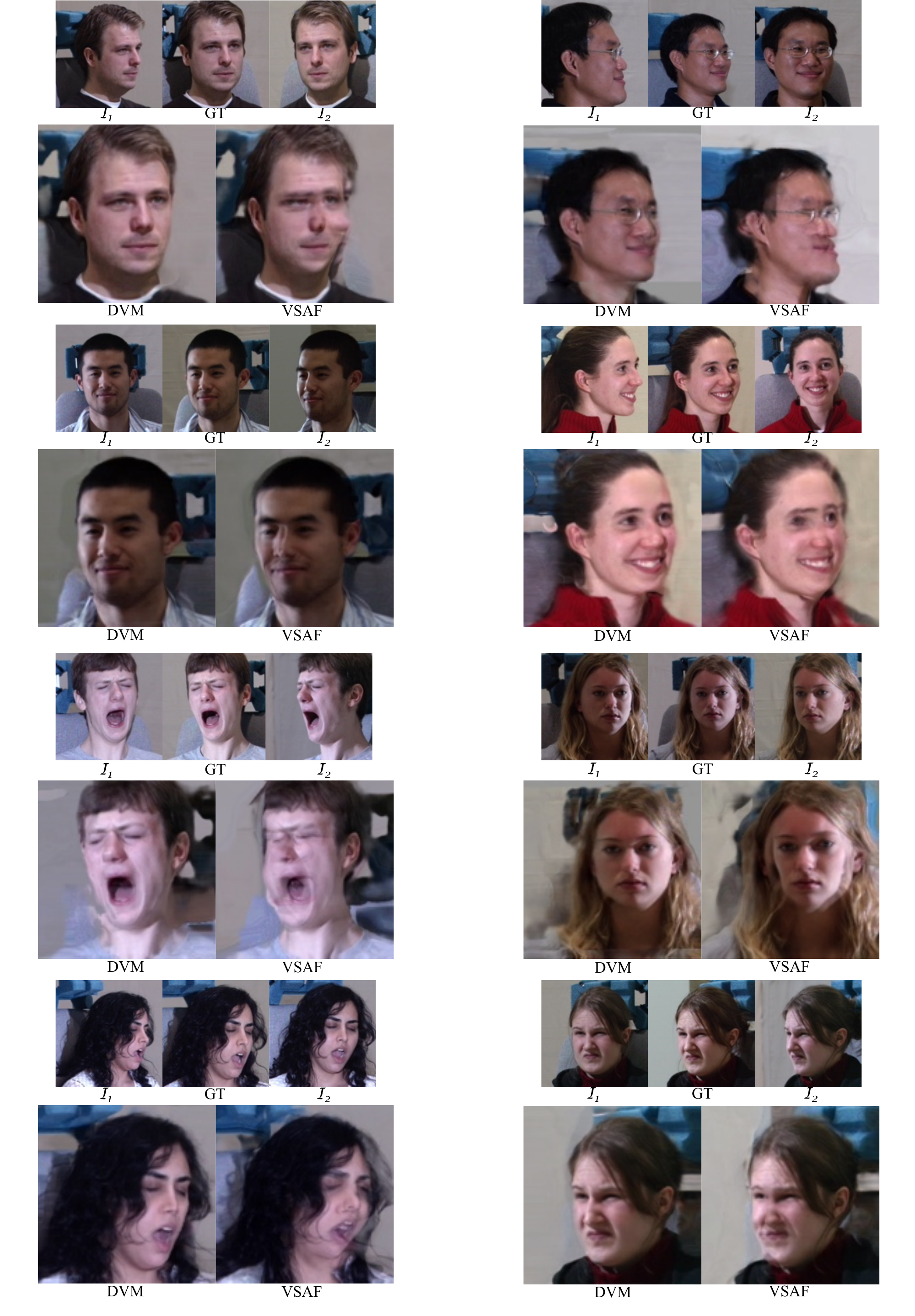}
\caption{Comparisons of view synthesis results by DVM and VSAF on test samples of Multi-PIE with the loose facial region crops. Two input images are shown on the left and right sides of the ground truth image (``GT").}
\label{fig:results_loose_face}
\end{figure*}

\begin{figure*}
\centering
\includegraphics[width=6in]{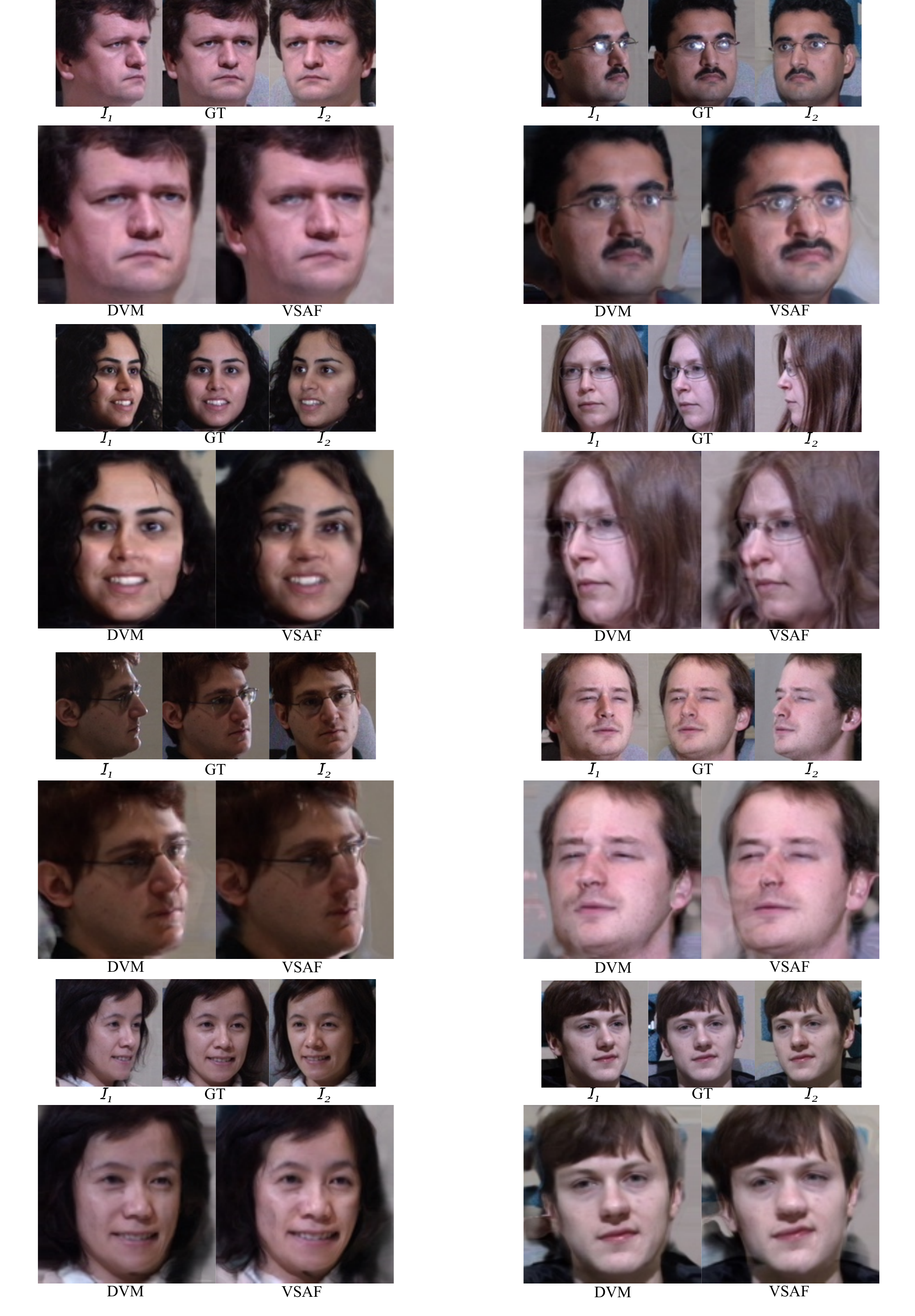}
\caption{Comparisons of view synthesis results by DVM and VSAF on test samples of Multi-PIE with the tight facial region crops. Two input images are shown on the left and right sides of the ground truth image (``GT").}
\label{fig:results_tight_face}
\end{figure*}

\begin{figure*}
\centering
\includegraphics[width=4.3in]{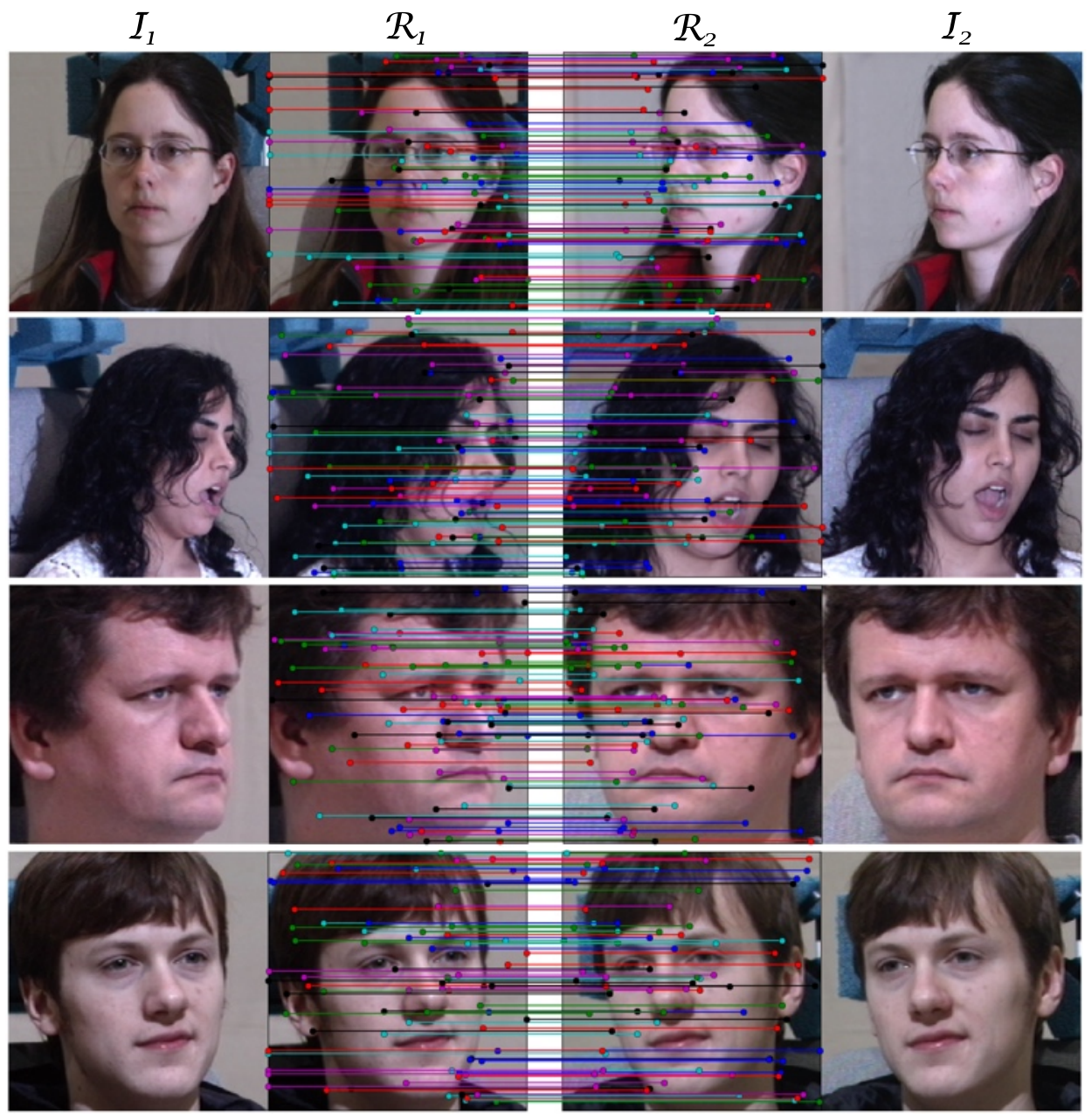}
\caption{Examples of rectification results and dense correspondences obtained by DVM on the Multi-PIE test input images.}
\label{fig:corr_face}
\end{figure*}

Figure~\ref{fig:results_loose_face} and Fig.~\ref{fig:results_tight_face} show more qualitative comparisons of the view synthesis results by DVM and VSAF on the Multi-PIE test data with the loose and tight facial region crops, respectively. Figure~\ref{fig:corr_face} shows examples of the rectification results and dense correspondence results by DVM on the Multi-PIE test input images.\\

\subsection*{C.3. Challenging cases}

\begin{figure*}
\centering
\includegraphics[width=6in]{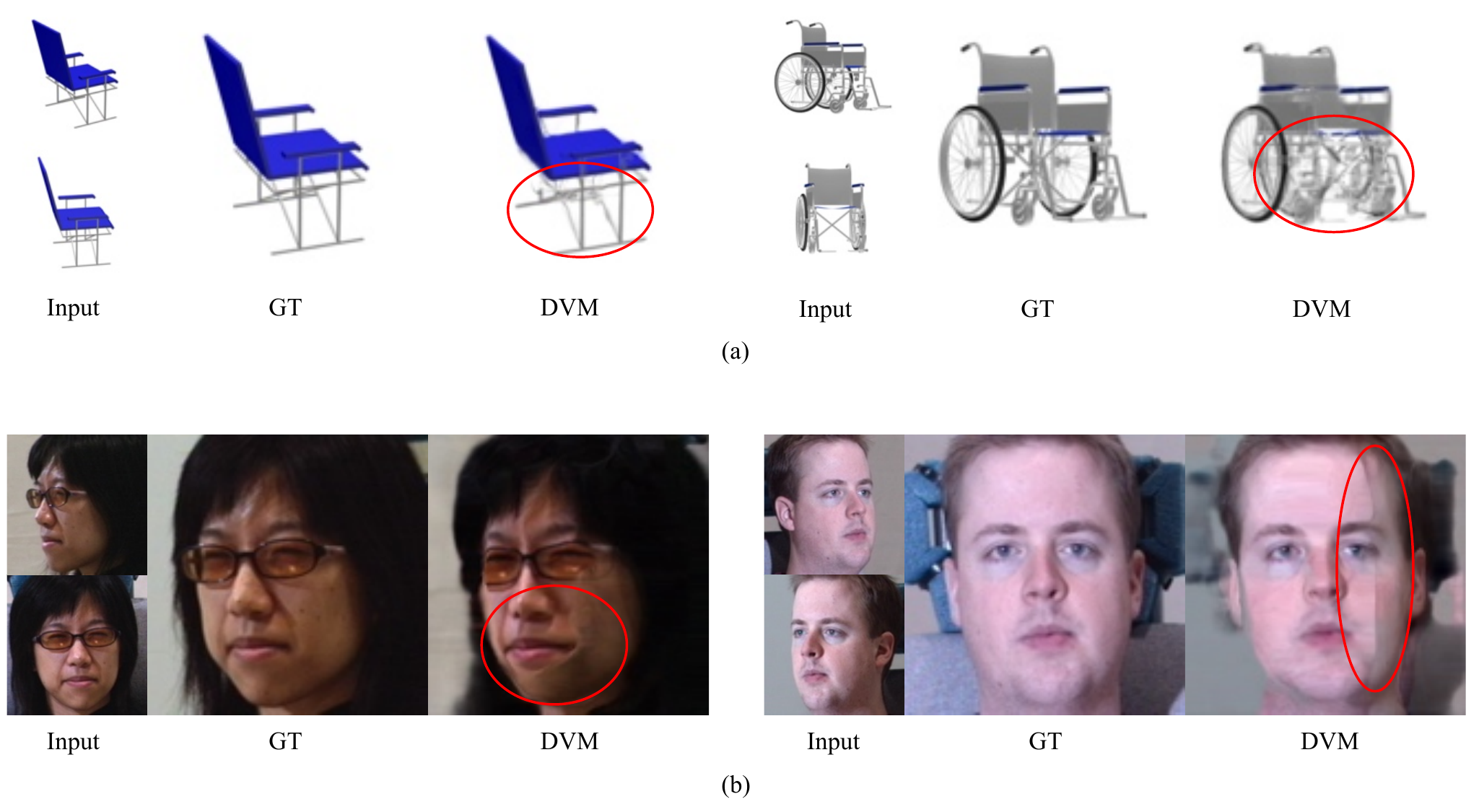}
\caption{Examples of challenging cases for Deep View Morphing.}
\label{fig:difficult_cases}
\end{figure*}

Figure~\ref{fig:difficult_cases} shows examples of challenging cases for DVM. It is generally difficult for DVM to deal with highly complex thin structures as shown in Fig.~\ref{fig:difficult_cases}(a). Plus, the current blending masks cannot properly deal with the different illumination and color characteristics between input images, and thus blending seams can be visible in some cases as shown in Fig.~\ref{fig:difficult_cases}(b).

\subsection*{C.4. Intermediate view synthesis}

We synthesize $\mathcal{R}_\alpha$ for any $\alpha$ of between 0 and 1 as 
\begin{equation}
\mathcal{R}_\alpha((1-\alpha)p_1^i + \alpha p_2^i)  =  w_1  (1 - \alpha) \mathcal{R}_1(p_1^i) + w_2  \alpha \mathcal{R}_2(p_2^i),
\label{eqn:intermediate_view}
\end{equation}
where $w_1 = \frac{(1-\alpha)\mathcal{M}_1(p_1^i)}{(1-\alpha)\mathcal{M}_1(p_1^i) + \alpha \mathcal{M}_2(p_2^i)}$ and $w_2 = 1 - w_1$. Note that we interpolate the blending masks $\mathcal{M}_1$ and $\mathcal{M}_2$ as well as $\mathcal{R}_1(P_1)$ and $\mathcal{R}_2(P_2)$. 

These synthesized views represent intermediate views between $\mathcal{R}_1$ and $\mathcal{R}_2$. As what we want in practice is intermediate views between $\mathcal{I}_1$ to $\mathcal{I}_2$, it is necessary to apply post-warping accordingly. We specifically create two linear camera paths, one from $\mathcal{I}_1$ to $\mathcal{R}_{\alpha=0.5}$ and the other from $\mathcal{R}_{\alpha=0.5}$ to $\mathcal{I}_2$. We can represent $H_1$ and $H_2$ used for rectifying $\mathcal{I}_1$ and $\mathcal{I}_2$ as elements of the special linear group $SL(3)$ by normalizing them to have unit determinants \cite{Kwon_PAMI}. Then the necessary post-warping homographies $H_\alpha$ for the linear camera paths can be determined as
\begin{equation}
H_\alpha = \left\{ 
\begin{matrix} 
\exp \left(  (1 - 2\alpha) \log \left( H_1^{-1}\right) \right), & \textrm{for } 0 < \alpha \leq 0.5 \\
\exp \left( (2\alpha - 1) \log \left( H_2^{-1} \right) \right), & \textrm{for } 0.5 < \alpha <1\\
\end{matrix}
\right. ,
\label{eqn:postwarping}
\end{equation}
where $\exp$ and $\log$ are matrix exponential and logarithm.

\begin{figure*}
\centering
\includegraphics[width=6in]{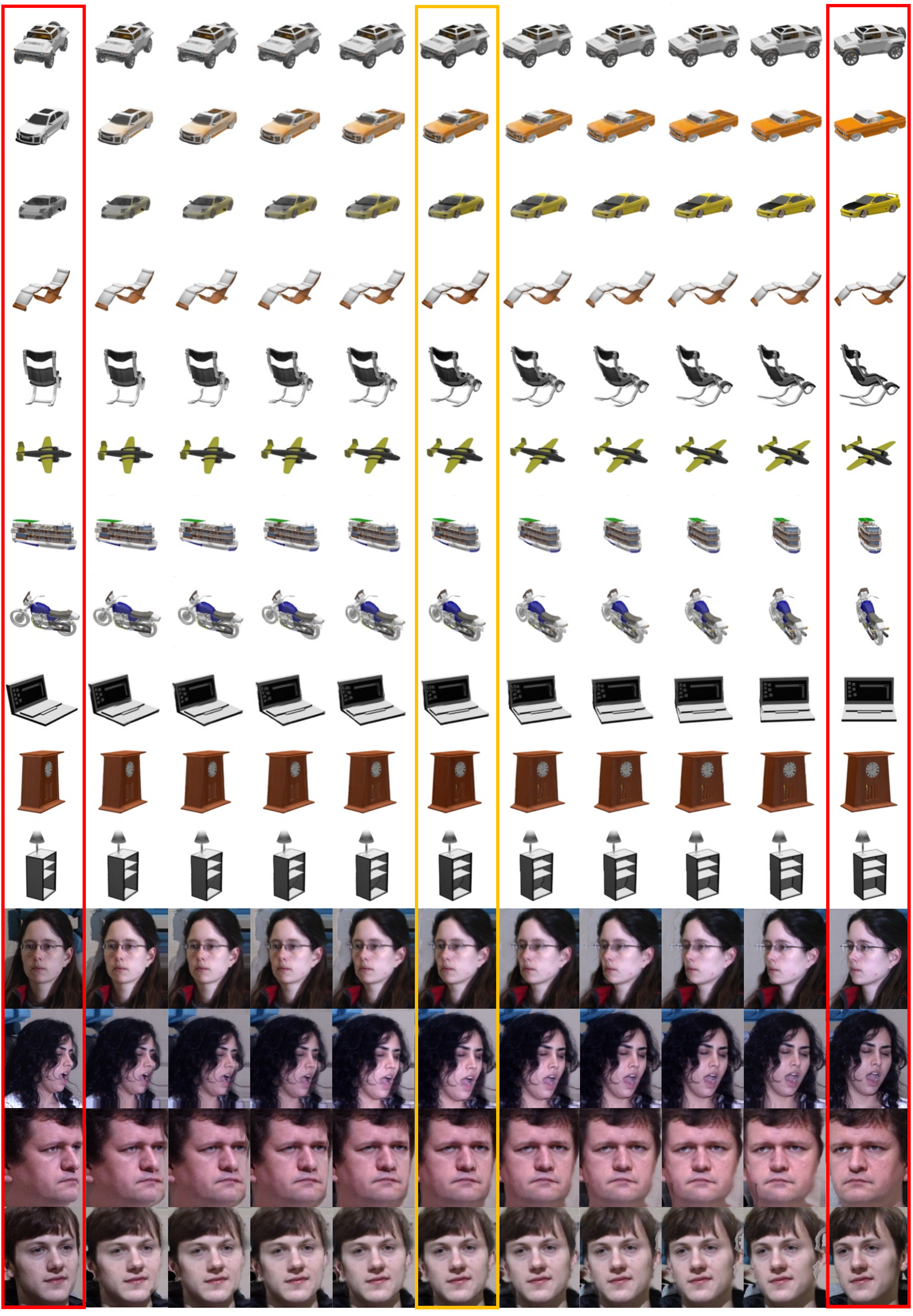}
\caption{Intermediate view synthesis results on the ShapeNet and Multi-PIE test input images. Red and orange boxes represent input image pairs and $\mathcal{R}_{\alpha=0.5}$ directly generated by DVM, respectively.}
\label{fig:interpolation_results_all}
\end{figure*}

Figure~\ref{fig:interpolation_results_all} shows the intermediate view synthesis results obtained by (\ref{eqn:intermediate_view}) and (\ref{eqn:postwarping}). The second and third rows of Fig.~\ref{fig:interpolation_results_all} show the intermediate view synthesis results for the case of input with different instances of ``Car", which are quite satisfactory. 

\end{document}